\pdfoutput=1

\documentclass[journal]{IEEEtran}
\usepackage{times}
\usepackage{epsfig}
\usepackage{graphicx}
\usepackage{amsmath}
\usepackage{amssymb}
\usepackage{pbox}
\usepackage{multirow}
\usepackage{subfig}
\usepackage[flushleft]{threeparttable}
\usepackage{color}

\newcommand{\specialcell}[2][c]{%
  \begin{tabular}[#1]{@{}c@{}}#2\end{tabular}}

\ifCLASSINFOpdf
\else
\fi

\begin{document}
\title{Frankenstein: Learning Deep Face Representations using Small Data}
\author{Guosheng~Hu,~\IEEEmembership{Member,~IEEE,}
        Xiaojiang~Peng, 
	Yongxin Yang, 
	Timothy M. Hospedales, 
        and~Jakob Verbeek
\thanks{G. Hu and J. Verbeek are with Univ.\ Grenoble Alpes, Inria, CNRS, Grenoble INP, LJK, 38000 Grenoble, France. email: \{guosheng.hu, jakob.verbeek\}@inria.fr.}
\thanks{X. Peng, corresponding author, is Hengyang Normal University, Hunan Province, China, email: xiaojiangp@gmail.com}
\thanks{Y. Yang is with  Electronic Engineering and Computer Science, Queen Mary University of London, UK. email: yongxin.yang@qmul.ac.uk}
\thanks{T. Hospedales is with the University of Edinburgh, UK. email: t.hospedales@ed.ac.uk}
}

\maketitle

\begin{abstract}
Deep convolutional neural networks have recently proven extremely effective for difficult face recognition problems in uncontrolled settings.
To train such networks, very large training sets are needed with millions of labeled images.
For some applications, such as near-infrared (NIR) face recognition, such large training datasets are  not publicly available and difficult to collect. 
In this work, we propose a method to generate very large training datasets of synthetic images by compositing real face images in a given dataset. 
We show that this method enables to learn models from as few as 10,000 training images, which perform on par with models trained from 500,000 images.
Using our approach we also obtain state-of-the-art results on the CASIA NIR-VIS2.0 heterogeneous face recognition dataset. 
\end{abstract}

\begin{IEEEkeywords}
face recognition, deep learning, small training data
\end{IEEEkeywords}

\IEEEpeerreviewmaketitle


\section{Introduction}
In recent years, deep learning methods, and in particular convolutional neural networks (CNNs),
 have achieved considerable success in a range of computer vision applications including object recognition \cite{krizhevsky12nips}, object detection \cite{girshick14cvpr}, semantic segmentation \cite{pinheiro15cvpr}, action recognition \cite{simonyan14nips}, and face recognition \cite{schroff15cvpr}. 
The recent success of CNNs stems from the following facts: 
(i) big annotated training datasets are currently available for a variety of recognition problems to learn rich models with millions of free parameters; 
(ii) massively parallel GPU implementations greatly improve the training efficiency of CNNs;
and (iii) new effective CNN architectures are being proposed, such as the VGG-16/19 networks~\cite{simonyan15iclr}, inception networks~\cite{szegedy15cvpr}, and deep residual networks~\cite{he2015deep}. 

Good features are essential for object recognition, including face recognition. 
Conventional features include linear functions of the raw pixel values, including Eigenface (Principal Component Analysis)~\cite{eigenface}, Fisherface (Linear Discriminant Analysis) 
~\cite{belhumeur97pami}, and Laplacianface (Locality Preserving Projection)~\cite{he05pami}.
Such linear features were later replaced by hand-crafted local non-linear features, such as Local Binary Patterns~\cite{LBP}, Local Phase Quantisation (LPQ)~\cite{LPQ}, and Fisher vectors computed over dense SIFT descriptors \cite{FVF}. 
Note that the latter is an example of a feature that also involves unsupervised learning.
These traditional features achieve promising face recognition rates in constrained environments, as  represented for example in the CMU PIE dataset \cite{sim02fg}.
However, using these features face recognition  performance may degrade dramatically in uncontrolled environments, such as  represented in the Labeled Faces in the Wild (LFW) benchmark~\cite{huang07lfw}.
To improve the performance in such challenging settings, metric learning can be used, see  \cite{JB,guillaumin09iccv1,weinberger09jmlr}.
Metric learning  methods learn a (linear) transformation of the features that pulls the objects that have the same label closer together, while pushing the objects that have different labels apart. 

Although hand-crafted features and metric learning achieve promising performance for uncontrolled face recognition, 
it remains cumbersome to improve the design of hand-crafted local features (such as SIFT \cite{lowe04ijcv}) and their aggregation mechanisms (such as Fisher vectors \cite{sanchez13ijcv}). 
This is because the experimental evaluation results of the features cannot be automatically fed back to improve the robustness  to nuisance factors such as pose, illumination and expression. 
The major advantage of CNNs is that all processing layers, starting from the raw pixel-level input, have configurable parameters that can be learned from data. This obviates the need for manual feature design, and replaces it with supervised data-driven feature learning. 
Learning the large number of parameters in CNN models (millions of parameters are rather a rule than an exception) requires very large training datasets. 
For example, the CNNs which achieve state-of-the-art performance on the LFW benchmark are trained using datasets with millions of labeled faces: Facebook's  DeepFace~\cite{taigman14cvpr} and Google's FaceNet~\cite{schroff15cvpr} were trained using 4 million and 200 million training samples, respectively. 

For some recognition problems large supervised training datasets can be collected relatively easily. For example the CASIA Webface dataset of 500,000 face images was collected semi-automatically from IMDb \cite{WEBFACE}.
However, in many other cases collecting large datasets may be costly, and possibly problematic due to privacy regulation. 
For example, thermal infrared imaging is ideal for low-light nighttime and covert face recognition applications \cite{kong05cviu}, 
but it is not possible to collect millions of labeled training images from the internet for the thermal infrared domain.
The lack of large training datasets is an important bottleneck that prevents the use of deep learning methods in such cases, as the models will overfit dramatically when using small training  datasets~\cite{MyDLeval}. 

To address this issue,  the use of big synthetic  training datasets has been explored by a number of authors ~\cite{jaderberg2014syn,papon2015semantic,rogez16nips,rozantsev2015rd}. 
There are two important advantages of using synthetic data 
(i) one can generate as many training samples as desired, and 
(ii) it allows explicit control over the nuisance factors. For instance, we can synthesize face images of all desired viewpoints, whereas data collected from the internet might be mostly limited to near frontal views.
Data synthesis has successfully been applied to diverse recognition problems, including text recognition~\cite{jaderberg2014syn}, scene understanding~\cite{papon2015semantic}, and object detection~\cite{rozantsev2015rd}. 
Several recent works \cite{zhu2015face,feng15tip,masi2016we,hu2016face,hu2017efficient} proposed 3D-aided face synthesis techniques for facial landmark detection and face recognition in the wild. 


Data augmentation is another technique that is commonly used to reduce the data scarcity problem \cite{paulin14cvpr,simonyan15iclr}. 
This is similar to data synthesis, but more limited in that existing training images are transformed without 
affecting the semantic class label, e.g.\ by applying cropping, rotation, scaling, etc. 

The main contribution of this paper is a solution for training deep CNNs using small datasets. 
To achieve this, we propose a data synthesis technique to expand limited datasets to larger ones that are suitable to train powerful deep CNNs.
Specifically, we synthesize images of a `virtual' subject $c$ by compositing automatically detected face parts (eyes, nose, mouth) of two existing subjects $a$ and $b$ in the dataset in a fixed pattern.
Images for the new subject are generated by composing  a nose from an image of subject $a$ with a mouth of an image of subject $b$.
This is motivated by the observation that face recognition consists in finding the differences in the appearance and constellation of face parts among people. 
For a dataset with an equal number of faces per person, this method can increase a dataset of $n$ images to one with $n^2$ images when using only 2 face parts (we use 5 parts in practice).
A dataset like LFW can thus be expanded from a little over 10,000 images to a dataset of 100 million images.

Unlike  existing face synthesis methods which use 3D models \cite{zhu2015face,feng15tip,masi2016we,hu2016face,hu2017efficient}, 
our method is a pure 2D method  which is much easier to implement. 
In addition, our method works on different tasks from  \cite{zhu2015face,feng15tip,masi2016we}. 
Specifically, the methods \cite{zhu2015face,feng15tip} are used for facial landmark detection, while ours for face recognition. 
The approach \cite{masi2016we} assumes a relatively large training data (500,000 images) already exists, while we assume the training data is very small (10,000 images). 

We experimentally demonstrate that the synthesized large training datasets indeed significantly improve the  generalization capacity of CNNs. 
In our experiments, we  generate a training set of 1.5 million images using an initial labeled dataset of only 10,000 images. 
Using the synthetic data we improve the face verification rate from 78.97\% to 95.77\% on LFW. 
In addition, the proposed face synthesis is also used for NIR-VIS heterogeneous face recognition \cite{ouyang2014hfrSurvey} and improve the rank-1 face identification rate from 17.41\% to 85.05\%.
With the synthetic data, we achieve state-of-the-art performance on both (1) LFW under the ``unrestricted, label-free outside data'' protocol 
and (2) CASIA NIR-VIS 2.0 database under rank-1 face identification protocol.


\section{Related work}
\label{sec:related}

Our work relates to three research areas that we briefly review below:
face recognition using deep learning methods (Section ~\ref{sec:FRDL}), face data collection  (Section ~\ref{sec:FDC}), and data augmentation and synthesis methods  (Section ~\ref{sec:SD}).

\subsection {Face recognition using deep learning}
\label{sec:FRDL}

Since face recognition is a special case of object recognition, 
good architectures for general object recognition may carry over to face recognition. 
Schroff et al.\ ~\cite{schroff15cvpr} explored networks that are based on that of Zeiler \& Fergus ~\cite{zeiler14eccv} and inception networks~\cite{szegedy15cvpr}. 
DeepID3 ~\cite{deepid3}  uses aspects of both inception networks and the very deep VGG network~\cite{simonyan15iclr}.  
Parkhi et al.~\cite{parkhi15bmvc} use the very deep VGG network, while Yi et al.\ \cite{WEBFACE} use $3\times 3$ filters but fewer layers. 
Hu et al. ~\cite{huattribute} use facial attribute information to improve the face recognition performance. 

DeepFace~\cite{taigman14cvpr} uses a 3D model for pose normalization, by which all the faces are rotated to the frontal pose. 
In this way,  pose variations are removed from the faces.
Then an 8-layer CNN is trained using four million pose-normalized images. 

DeepID ~\cite{DEEPID}, DeepID2~\cite{DEEPID2}, DeepID2+~\cite{DEEPID2P} 
all train an ensemble of small CNNs. The input of one small CNN is an image patch cropped around a facial part (face, nose, mouth, etc.). The same idea is also used in~\cite{BaiduFace}.
DeepID uses only a classification-based loss to train the CNN, while DeepID2  includes an additional verification-based loss function.
To further improve the performance,  DeepID2+ adds losses to all the convolutional layers rather than the topmost layer only.

All the above methods train CNNs using large training datasets of 500,000 images or more.  
To the best of our knowledge, only~\cite{MyDLeval}  uses small datasets to train CNNs (only around 10,000 LFW images) and achieves significantly worse performance on the LFW benchmark:  
87\% vs 97\% or higher in ~\cite{schroff15cvpr,DEEPID2P,taigman14cvpr}. 
Clearly, sufficiently large training datasets are extremely important for learning deep face representations.

\subsection {Face dataset collection} \label{sec:FDC}

Since big data is important for learning a deep face representation, several research groups have collected large datasets with 90,000 up to 2.6 million labeled face images~\cite{ JB,megaface,parkhi15bmvc,sun13iccv1,WEBFACE}.
To achieve this, they collect face images from the internet, by querying for specific websites such as IMDb or general search engines for celebrity names. 
This data collection process is detailed in~\cite{parkhi15bmvc,WEBFACE}. 

Existing face data collection methods have, however, two main weaknesses. 
First, and most importantly, internet-based collection of large face datasets is limited to visible spectrum images, and is not applicable to collect e.g. infrared face images.
Second, the existing collection methods are expensive and time-consuming. 
It results from the fact that automatically collected face images are noisy, and manual filtering has to be performed to remove incorrectly labeled images~\cite{parkhi15bmvc}. 

The difficulty of collecting large datasets in some domains, e.g. for infrared imaging, motivates the work presented in this paper.
To address this issue we propose a data synthesis method that we describe in the next section. 

\subsection {Data augmentation and synthesis}
\label{sec:SD}

The availability of large supervised datasets is the key for machine learning to succeed, and this is true in particular for very powerful deep CNN models with millions of parameters.  
To alleviate data scarcity in visual recognition tasks, data augmentation has been used to add more examples by applying simple image transformations that do not affect the semantic-level image label, see e.g. \cite{decoste02ml}. Examples of such transformations are horizontal mirroring, cropping, small rotations, etc. 
Since it is not always clear in advance which (combinations of) transformations are the most effective to generate examples that improve the learning the most, Paulin et al.~\cite{paulin14cvpr} proposed to learn which transformations to exploit. 

Data augmentation, however, is limited to relatively simple image transformations. Out-of-plane rotations, for example, are hard to accomplish since they would require some degree of 3D scene understanding from a single image. Pose variations of articulated objects are another example of transformations that are non-trivial to obtain, and generally not used in data augmentation methods.

Training models from synthetic data can overcome such difficulties, provided that sufficiently accurate object models are available.
%
Recent examples where visual recognition systems have been trained from synthetic data include the following. 
Shotton et al.~\cite{shotton13acm} train randomized decision forests for human pose estimation from synthesized 3D depth data.
Jaderberg et al.~\cite{jaderberg2014syn} use  synthetic data to train CNN models for natural scene text recognition.
Su et al.~\cite{su15iccv} use synthetic images of objects to learn a CNN for viewpoint estimation. 
Papon and Schoeler~\cite{papon2015semantic}
train a multi-output CNN that predicts class, pose, and location of objects from realistic cluttered room scenes that are synthesized on the fly. 
Weinmann et al.~\cite{weinmann14eccv} synthesize material images under different viewing and lighting conditions based on detailed surface geometry measurements, and use these to train a recognition system using a SIFT-VLAD representation \cite{jegou10cvpr}.
Ronzantsev et al.~\cite{rozantsev15cviu} use rough 3D models to synthesize new views of real object category instances. 
They show that this outperforms more basic data augmentation using crops, flips, rotations, etc.

Data synthesis techniques are also used for face analysis. 
To improve the accuracy of facial landmark detection in the presence of large pose variations~\cite{feng15tip, zhu2015face}, a 
3D morphable face models is used to synthesize face images in arbitrary poses. 
Similar data synthesis techniques are also used for pose-robust face recognition \cite{masi2016we}.
Unlike 3D solutions, we propose a 2D data synthesis method to solve the problem of training deep CNNs using very limited training data.



\begin{figure} 
\begin{center}
\includegraphics[trim = 25mm 60mm 25mm 25mm, clip, width=0.65 \linewidth]{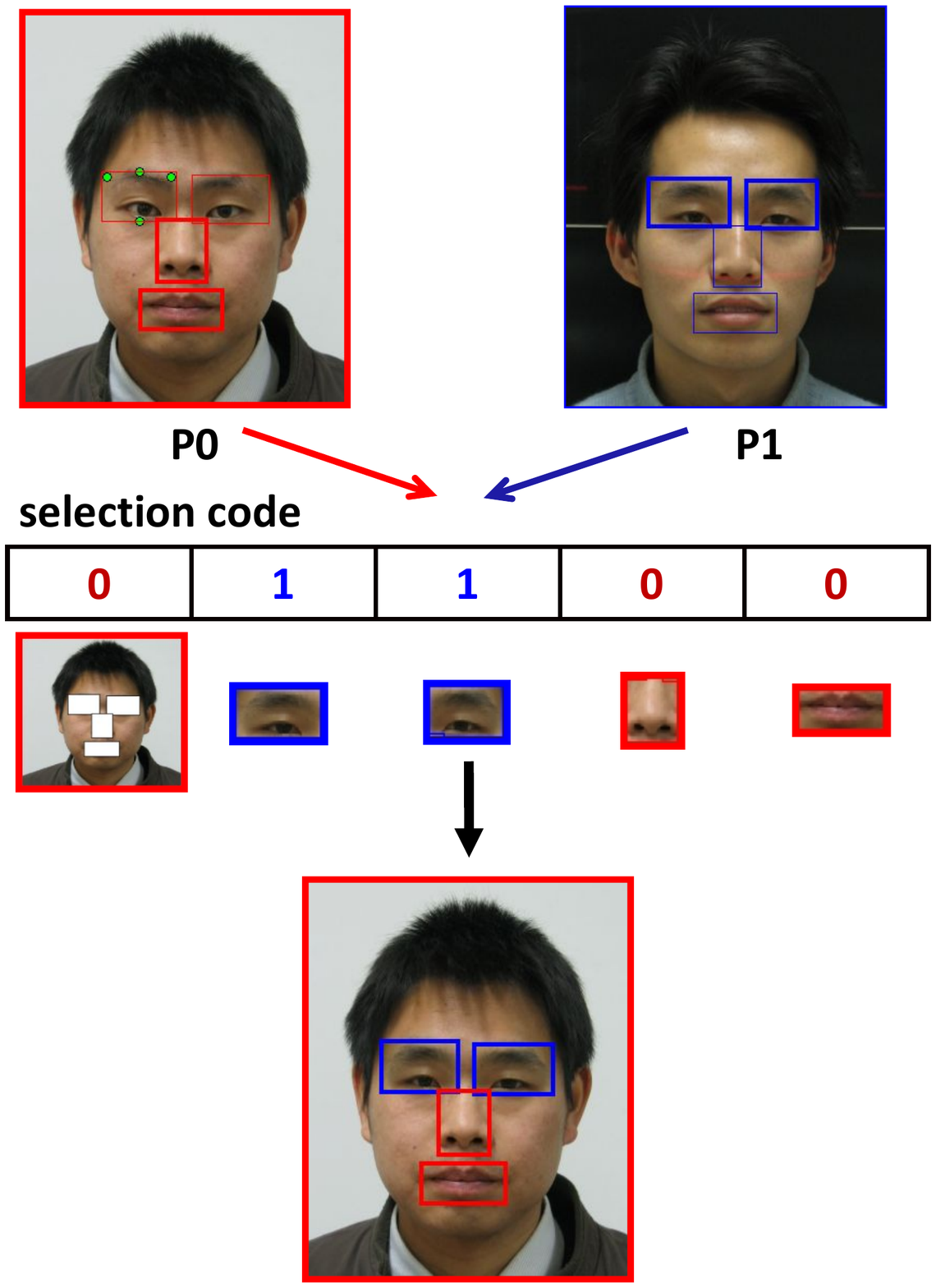}
\end{center}
\caption{Illustration of the face synthesis process using five parts: left-eye, right-eye, nose, mouth and the rest.
Parent images P0 and P1 (top) are mixed by using the eyes of P1 and the other parts of P0   to form the synthetic image (bottom). 
} 
\label{fig:engine}
\end{figure}

\section{Synthetic data engine}
\label{sec:SDE}

Human faces are well structured in the sense that they are composed of parts (eyes, nose, mouth, etc.) which are organized in a relatively rigid constellation. 
Face recognition relies on differences among people in the appearance of facial parts and their constellation.
Motivated by this, our synthetic face images are generated by swapping one or more facial parts among existing ``parent'' images. 
In our work we use five face parts: right eye (RE), left eye (LE), nose (N), mouth (M) and the rest (R). 
See Figure~\ref{fig:engine} for an illustration. 
 For simplicity, we only consider the synthesis using only two parent images  in this work. 
Our synthesis method can easily be extended, however, to the scenario of more than two parent images.

\subsection{Compositing face images}
Suppose that we have an original dataset and let  $\mathcal S$ denote the set subjects in the dataset, and let $n_i$ denote the number of images of subject $i\in \mathcal S$. 
To synthesize an image, we select a tuple $(i,j,c,s,t)$ where $i\in {\mathcal S},j\in {\mathcal S}$ correspond to two subjects that will be mixed, and $s\in \{1,\dots, n_i\}$ and $t\in\{1,\dots,n_j\}$ are indices of images of $i$ and $j$ that will be used.
The bitcode $c\in\{0,1\}^5$ defines which parts will be taken from each subject.
A zero at a certain position in $c$ means that the corresponding part will be taken from $i$, otherwise it will be taken from $j$.
There are only $2^5-2=30$ valid options for $b$, since the codes 00000 and 11111 correspond to the original images of $s$ and $t$ respectively, instead of synthetic ones. 

To synthesize a new image, we designate one of  parent images as the ``base'' image from which we use the R (the rest) part, and the other as the ``injection'' image from which one or more parts will be pasted on the base image.
Since the size of the  facial parts in the two parent images are in general different,
 we re-size the facial parts of the injection image to that of the base image. 
The main challenge to implement the proposed synthesis method is to accurately locate the facial parts. 
Recently, many efficient and accurate landmark detectors have been proposed. 
We use four landmarks detected by the method of Zhang et al.~\cite{zhang14eccv} to define the rectangular region that corresponds to each face part.

We refer to each choice of $(i,j,c)$ with $i\neq j$ as a ``virtual subject'' which consists of a mix of two existing subjects in the dataset. In total we can generate $30 |{\mathcal S}| (|{\mathcal S}|-1)/2$ different virtual subjects, and for each of these we can generate $n_i\times n_j$ samples. 
Note that if we set $i=j$ we can in the same manner synthesize 
$30 n_i (n_i-1)/2 $ new images for an existing subject.

Although some works \cite{masi2016we, deepface} empirically verified the effectiveness of synthetic data, they did not give much insight into how. 
In our work, the synthetic data captures a dataset of richer intra-personal variations by generating a large number of images of the same identities, leading to a `deeper' training set. 
Also, our engine can synthesize a large number of faces of new identities, generating a `wider' training set. Thus the synthetic identities interpolate the whole space of pixel-identity mappings. 
Not surprisingly, a better CNN model can be trained using this deeper and wider training set. 
The methods of generating our deeper and wider training set are detailed in Section ~\ref{facialdatasynt}. 

\subsection{Compositing artefacts} 
\begin{figure} 
\begin{center}
\includegraphics[trim = 35mm 48mm 5mm 40mm, clip, width=1 \linewidth]{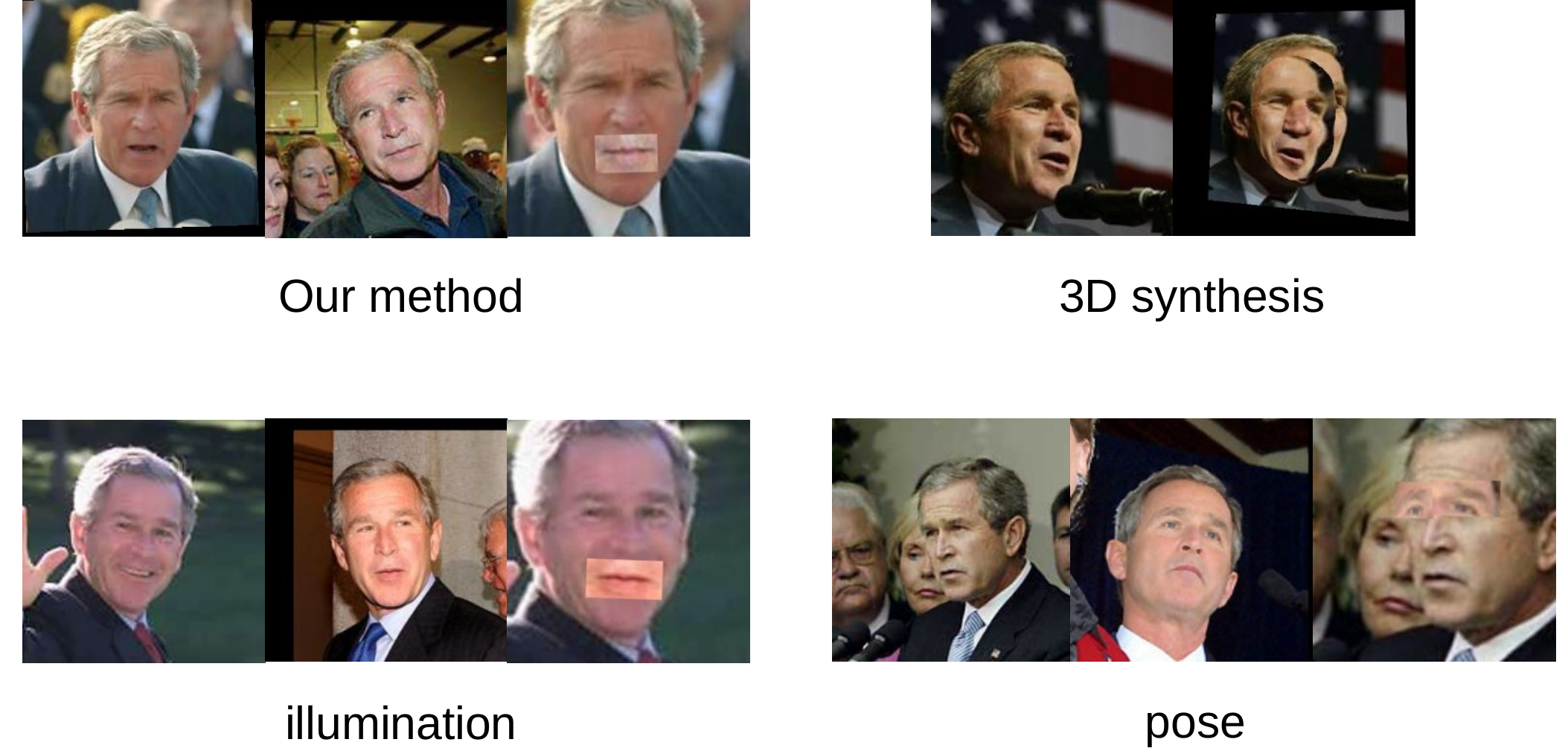}
\end{center}
\caption{
Top row: Type I hard boundary artefacts generated by our method (left) and 3D synthesis methods~\cite{masi2016we, deepface, hassner2014effective} (right).
Bottom row: Type II artefacts due to inconsistencies in illumination (left) and pose (right) generated by our method. 
For our method we show the two parent images followed by the composited image.
} 
\label{fig:artefacts}
\end{figure}

The synthetic faces present two types of artefacts: (I) hard boundary effects, and  
(II)  inconsistent/unnatural intra-personal variations (lighting, pose, etc.) between facial patches.
These are illustrated in Fig.~\ref{fig:artefacts}. 
Note that the type~I artefacts are generated by not only our method but also 3D synthesis methods such as \cite{masi2016we, deepface, hassner2014effective}. 
As shown in the top-right side of Fig.~\ref{fig:artefacts}, the artefacts created by 3D methods are due to inaccurate 3D model to 2D image fitting. 
The inaccurate fitting makes the synthetic faces extract the pixels from background rather than facial areas, leading to bad facial boundaries.

Despite the existence of these artefacts, this synthetic data is still useful for training strong face recognition models. This can be understood from several perspectives: (1) Type~I artefacts are common to all the synthetic faces in the training set, therefore the CNN does not learn to rely on artefacts as discriminative features coding for identity, i.e., it learns artefact invariance. This means its performance is not compromised when subsequently presented with artefact-free images at testing-time. Other studies have also shown that synthetic data still improves recognition performance, despite the presence of type~I artefacts ~\cite{masi2016we}.
(2) The artefacts can be regarded as noise, which has been shown to improve model generalisation in a variety of settings by increasing robustness and reducing overfitting. For example in the case of CNNs, training with data augmentation in the form of specifically designed deformation noise is important to obtain good recognition performance \cite{yu2016sketchAnet}; and in the case of de-noising auto encoders, training on images with noise, corruption and artefacts has been shown to improve face classification performance~\cite{vincent2010stacked}. (3) As a concrete example to understand how type II artefacts can improve performance, consider two synthetic images with the same identity label, but one a type II artefact on the mouth caused by illumination, e.g., in Fig.~\ref{fig:artefacts} (bottom left). Training to predict these images as having the same identity means that the CNN learns an illumination-invariant feature for the mouth. And similarly for other intra-personal variations (such as pose, expression). Thus while some artefact images look strange, they are actually a powerful form of data augmentation that helps the CNN to learn robustness to all these nuisance factors.

\section{Face recognition pipeline}
\label{sec:method}

In this section we describe the different elements of our pipeline for face identification and verification in detail.

\subsection{CNN architectures}
\label{sec:MFritw}

Face recognition in the wild is a challenging task. 
As described in Section ~\ref{sec:FRDL}, the existing deep learning methods highly depend on big training data.
Very little research investigates training CNNs using small data.
Recently, Hu et al.\ ~\cite{MyDLeval} evaluated CNNs trained using small datasets.
Due to the limited training samples,  they found the  performance of CNNs to be  worse than handcrafted features such as high-dimensional features~\cite{HimF} 
 (0.8763 vs 0.9318).
In this work, we  use a limited training set of around 10,000 images to synthesize a much larger one of around 1.5 million images  for CNN training. 
The synthesized  training data captures various deformable facial patterns that are important to improve the generalization capacity of CNNs. 

We use two CNN architectures.
The first one, from~\cite{MyDLeval}  has fewer filters and is referred as CNN-S. 
The second, from~\cite{WEBFACE},  is much larger and  referred as CNN-L.
These two architectures are detailed in Table~\ref{tab:2CNNs}.
Using  the CNN-L model we achieve state-of-the-art performance on the LFW dataset \cite{huang07lfw} under  `unrestricted, label-free outside data' protocol.

\begin{table}
\centering
\caption{ {Our two CNN architectures}}
\label{tab:2CNNs}
\begin{tabular}{|c|c|}
\hline
\multicolumn{1}{|c|}{CNN-L}                                  & \multicolumn{1}{c|}{CNN-S} 	\\ \hline
\multicolumn{2}{|c|}{conv1}                                                     \\ \hline
 \specialcell{32$\times$3$\times$3, st.1;  64 $\times$ 3 $\times$ 3, st.1 \\x2 maxpool, st.2} & \specialcell{16$\times$3$\times$3, st.1;  16 $\times$ 3 $\times$ 3, st.1\\ x2 maxpool, st.2}  		\\ \hline
 \multicolumn{2}{|c|}{conv2}                                                     	\\ \hline
    \specialcell{64$\times$3$\times$3, st.1;  128 $\times$ 3 $\times$ 3, st.1 \\x2 maxpool, st.2}                   & \specialcell{32$\times$3$\times$3, st.1 \\x2 maxpool, st.2}                     \\ \hline
\multicolumn{2}{|c|}{conv3}                                                     	\\ \hline
     \specialcell{96$\times$3$\times$3, st.1 ; 192 $\times$ 3 $\times$ 3, st.1 \\x2 maxpool, st.2}                  & \specialcell{48$\times$3$\times$3, st.1 \\x2 maxpool, st.2}                     \\ \hline
\multicolumn{2}{|c|}{conv4}                                                     	\\ \hline
      \specialcell{128$\times$3$\times$3, st.1 ; 256 $\times$ 3 $\times$ 3, st.1 \\x2 maxpool, st.2}                  &        -              \\ \hline
\multicolumn{2}{|c|}{conv5}    \\ \hline
     \specialcell{160$\times$3$\times$3, st.1 ; 320 $\times$ 3 $\times$ 3, st.1 \\x7 avgpool, st.1}                    &         -             \\ \hline 
\multicolumn{2}{|c|}{fully connected}    \\ \hline
     \specialcell{Softmax-5000}                    &         \specialcell{FC-160 \\ Softmax-5000}              \\ \hline 
\end{tabular}
\end{table}

\subsection{NIR-VIS heterogeneous face recognition}
 NIR-VIS (near-infrared to visual) face recognition is important in applications where probe images are captured by NIR cameras that use active lighting which is invisible to the human eye \cite{ouyang2014hfrSurvey}. 
 Gallery images are, however, generally only available in the visible spectrum.
 The existing methods for NIR-VIS face recognition include three steps: (i) illumination pre-processing, (ii) feature extraction, and (iii) metric learning. 
 First, the  NIR-VIS illumination differences cause the main difficulty of NIR-VIS face recognition. 
 Therefore, illumination normalization methods are usually used to reduce these differences. 
 Second, to reduce the heterogeneities of NIR and VIS images, illumination-robust features such as LBP  are usually extracted. 
 Third, metric learning is widely utilized, aiming at removing the differences of modalities and meanwhile keeping the discriminative information of the extracted features. 
 

In this work, we also follow these three steps that are detailed in Section~\ref{sec:ExpNirVis}. 
Unlike the existing work that extracts handcrafted features, we learn face representations using  two CNN architectures described above.  
To  our knowledge, we are the first to use deep CNNs for NIR-VIS face recognition.
The main difficulty of training CNNs results from the lack of NIR training  images, which we address via data synthesis.

\subsection{Network Fusion}
\label{sec:NFusion}

Fusion of multiple networks is a widely used strategy to improve the performance of deep CNN models. 
For example, in~\cite{simonyan15iclr} an ensemble of seven networks is used to improve the object recognition performance due to complementarity of the models trained at different scales.
 Network fusion is also successfully applied to learn face representations. 
 DeepID and its variants~\cite{DEEPID2,DEEPID,DEEPID2P} train multiple CNNs using image patches extracted from different facial parts. 

The heterogeneity of NIR and VIS images is intrinsically caused by the different spectral bands from which they are acquired. 
The images in both modalities, however, are reflective in nature and affected by illumination variations.
Illumination normalization can be used to reduce such variability, at the risk of loosing identity-specific characteristics.
In this work, we fuse two networks that are trained using the original  and illumination-normalized images respectively.
This network fusion 
significantly  boosts the recognition rate.

\subsection{Metric Learning}
\label{eq:mln} 

The goal of metric learning is to make different classes more separated,  and instances in the same class closer. 
Most approaches learn a  Mahalanobis metric
\begin{equation}
d^2_A(x_i,x_j) = (x_i-x_j)^T A (x_i-x_j),
\label{Mah}
\end{equation} 
 which maximizes inter-class discrepancy, while minimizing intra-class discrepancy.
Other methods, instead learn a generalized dot-product of the form 
\begin{equation}
d^2_B(x_i,x_j) = x_i^T B x_j.
\label{Cos}
\end{equation}

Metric learning methods are widely used for face identification and verification. 
Because  identification and verification are two different tasks, different loss functions should be optimized to learn the metric.
Joint Bayesian metric learning (JB)~\cite{JB} and Fisher linear discriminant analysis (LDA) are probably the two most widely used metric learning methods for face verification and identification respectively. 
In particular, LDA can be seen as a method to learn a metric of the form of Eq.~(\ref{Mah}), while JB learns a verification function that can be written as a weighted sum of Eq.~(\ref{Mah}) and (\ref{Cos}).
In our work we use  JB and LDA to improve the performance of face verification and identification respectively.
 

\section{Experiments}
\label{sec:EXP}

\subsection{Data synthesis methods}
\label{facialdatasynt}
\begin{figure*}[t]
\begin{center}
\includegraphics[ width=0.3 \linewidth]{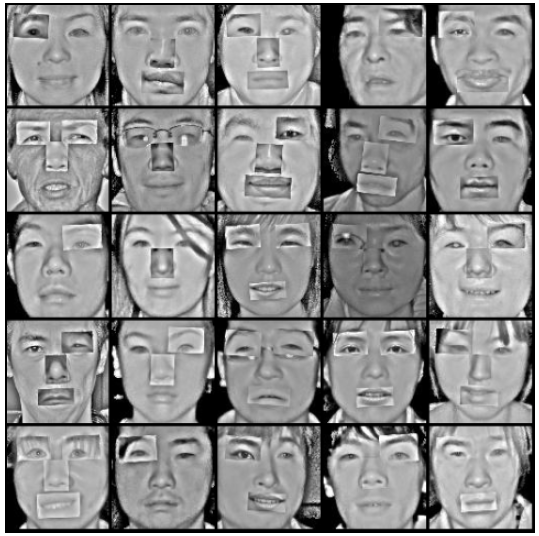}
\includegraphics[ width=0.3 \linewidth]{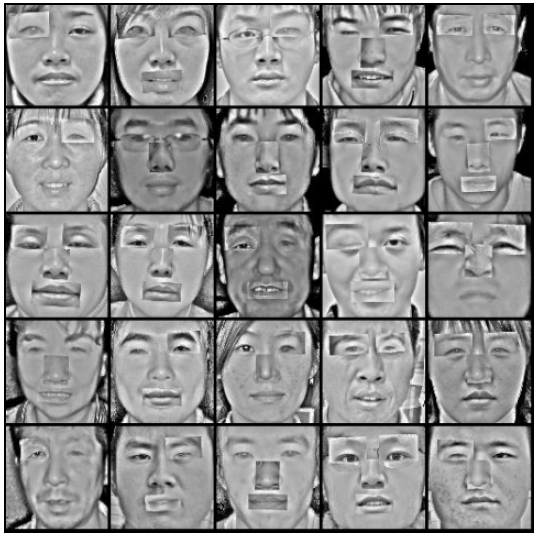}
\includegraphics[ width=0.3 \linewidth]{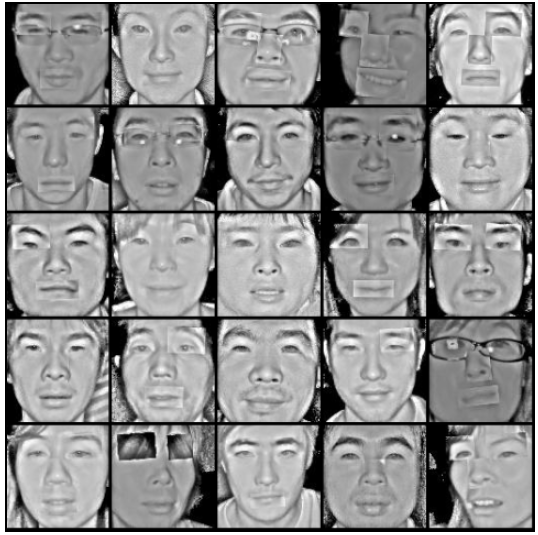}
\end{center}
\caption{ Left: Inter-Synthesis. Middle: cross-modality Intra-Synthesis. Right: Intra-Synthesis.}
\label{fig:syn}
\end{figure*}

Given a set of face images and their IDs, we define three strategies for synthesis: 
\emph{Inter-Synthesis}, \emph{Intra-Synthesis}, and \emph{Self-Synthesis}.
\emph{Inter-Synthesis} synthesizes a new image using two parents from different IDs as shown in Fig.\ 1.
The facial components of an  \emph{Intra-Synthesized} face are from different images with the same ID.
\emph{Self-synthesis} is a special case of \emph{Intra-Synthesis}. Specifically, one given image  synthesizes new images by swapping facial components of itself and its mirrored version.
By virtue of \emph{Self-Synthesis}, one input image can become maximum 32 images which have complementary information.
In the view of  NIR-VIS  cross-modality, we also define `cross-modality synthesis' which uses images from different modalities to synthesize a new one.
Some synthetic images from the CASIA NIR-VIS 2.0 dataset with LSSF~\cite{LSSF} illumination normalization are shown  in Fig. \ref{fig:syn}.
The reasons of using LSSF illumination normalization are detailed in Section~\ref{sec:ExpNirVis}.
As shown in Fig. \ref{fig:syn}, the results of \emph{Intra-Synthesis} method are usually more natural than \emph{Inter-Synthesis} method since the \emph{Intra-Synthesis} method uses the same ID.
However, as shown in the right of Fig. \ref{fig:syn}, 
\emph{Intra-Synthesis} can also create artefacts due to large pose variations.

\subsection{Implementation details}
\label{sec:Impdetails}

Before face synthesis, all the raw images are aligned  and cropped to size $100 \times 100$ as in ~\cite{WEBFACE} on both datasets.
We train our models using images only from LFW and CASIA NIR-VIS2.0 databases.
For the CNN-S model on both datasets, we set the learning rate as 0.001, and decrease it by 10 times every 4000 iterations, and stop training after 10K iterations. 
We find that dropout is not helpful for the small network, and report results obtained without it.
For the CNN-L model on the NIR-VIS dataset, we set the learning rate as 0.01, and decrease it by 10 times every 8000 iterations, and stop training after 20K iterations.
For the CNN-L model on the LFW dataset, we set the learning rate as 0.01, and decrease it by 10 times every 120K iterations, and stop training after 200K iterations.
We set dropout rate as 0.4 for the pool5 layer of the CNN-L model. For both CNN-S and CNN-L models, the batch size is 128, momentum is 0.9, and decay is 0.0005.
Softmax loss function is used to guide CNN training.
The features used in our recognition experiments with CNN-S and CNN-L are FC-160 (160D) and Pool5 (320D), respectively.

\subsection{Face recognition in the wild}
\label{sec:ExpFRW}
\subsubsection{Database and protocol} 
Labeled Faces in the Wild (LFW) ~\cite{huang07lfw} is a public dataset for unconstrained face recognition study. It contains 5,749 unique identities and 13,233 face photographs.
The training and test sets are pre-defined in~\cite{huang07lfw}.
For evaluation, the full dataset is divided into ten splits, and each time nine of them are used for training and the left one for testing.
Our work falls in the protocol of `Unrestricted, Label-Free Outside Data' as we use the identity information to train the neural network (softmax loss). Meanwhile, all face images are aligned using a model trained on unlabeled outside data.
As a benchmark for comparison, we report the mean and standard deviation of classification accuracy.

Under LFW protocol, the training set in each fold is different.
Therefore, the size of synthetic data and the original raw LFW data in Table~\ref{tab:SDGLFW} is averaged over 10 folds.
We generate 1.5 million training images, including 1 million `Inter-Syn' ones and 0.5 million `Intra-Syn' ones, as defined in Section~\ref{facialdatasynt}.

\begin{table}
\centering
\caption{Training data synthesized from LFW}
\label{tab:SDGLFW}
\begin{tabular}{|c|c|c|c|c|}
\hline
	  &   			& IDs & Images & Images/ID \\ \hline 
\multirow{3}{*}{Synthetic}	 &Intra-Syn & 5K       & 500K   & 100           \\ \cline{2-5}
			  	&Inter-Syn & 5K       & 1M     & 200           \\ \cline{2-5}
			   	&Total     & 10K      & 1.5M   & 150           \\ \hline
\multicolumn{2}{|c|}{Raw}	&5K          & 10K       & 2 \\ \hline
\end{tabular}
\end{table}

\begin{figure}
\begin{center}
\includegraphics[trim = 40mm 60mm 40mm 25mm, clip, width=1 \linewidth]{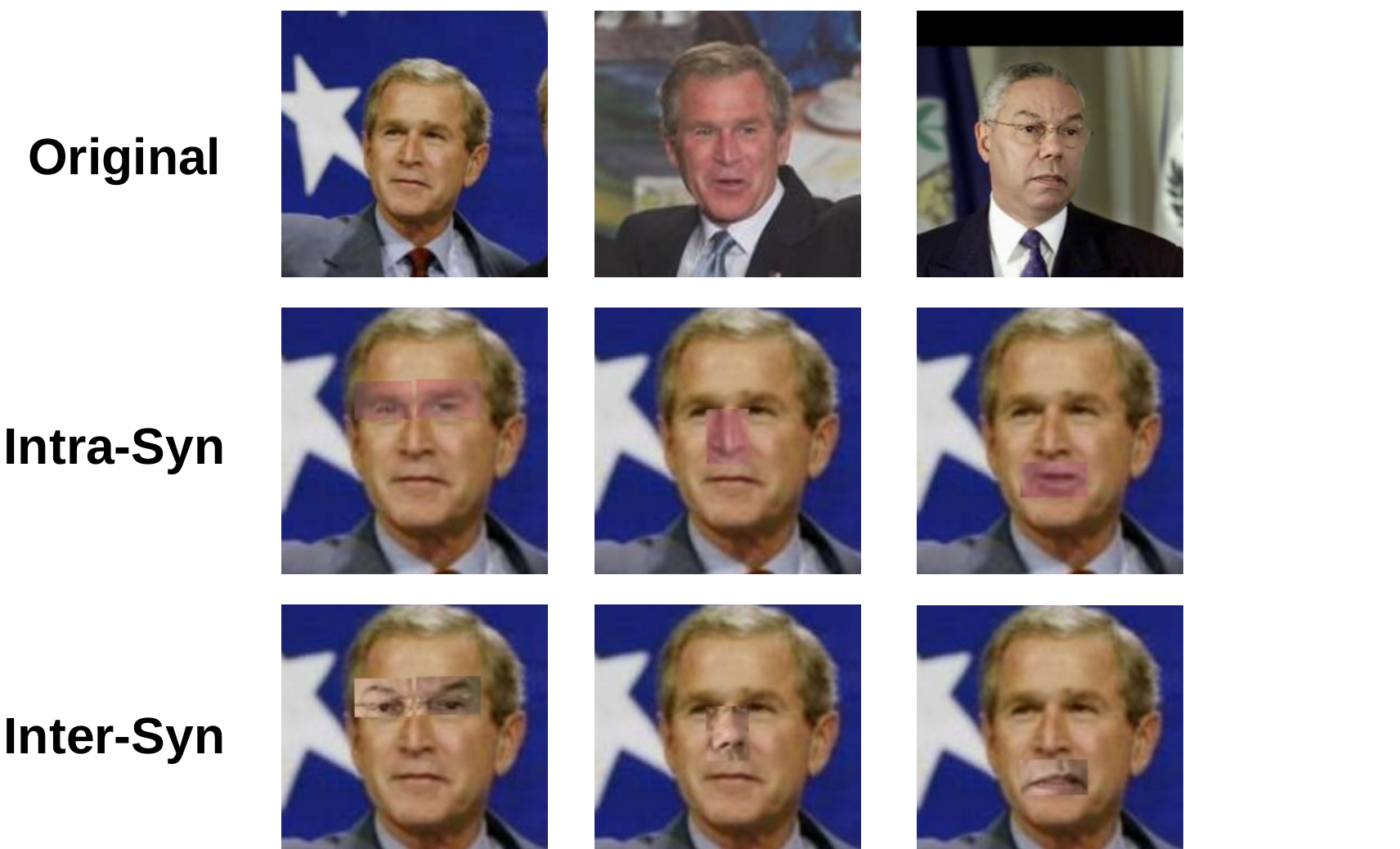}
\end{center}
\caption{Samples of Inter-Synthesis and Intra-Synthesis} 
\label{fig:vissyn}
\end{figure}

\begin{figure*}
\begin{center}
\subfloat[]{
\includegraphics[trim=60 180 60 230,width=0.32\textwidth]{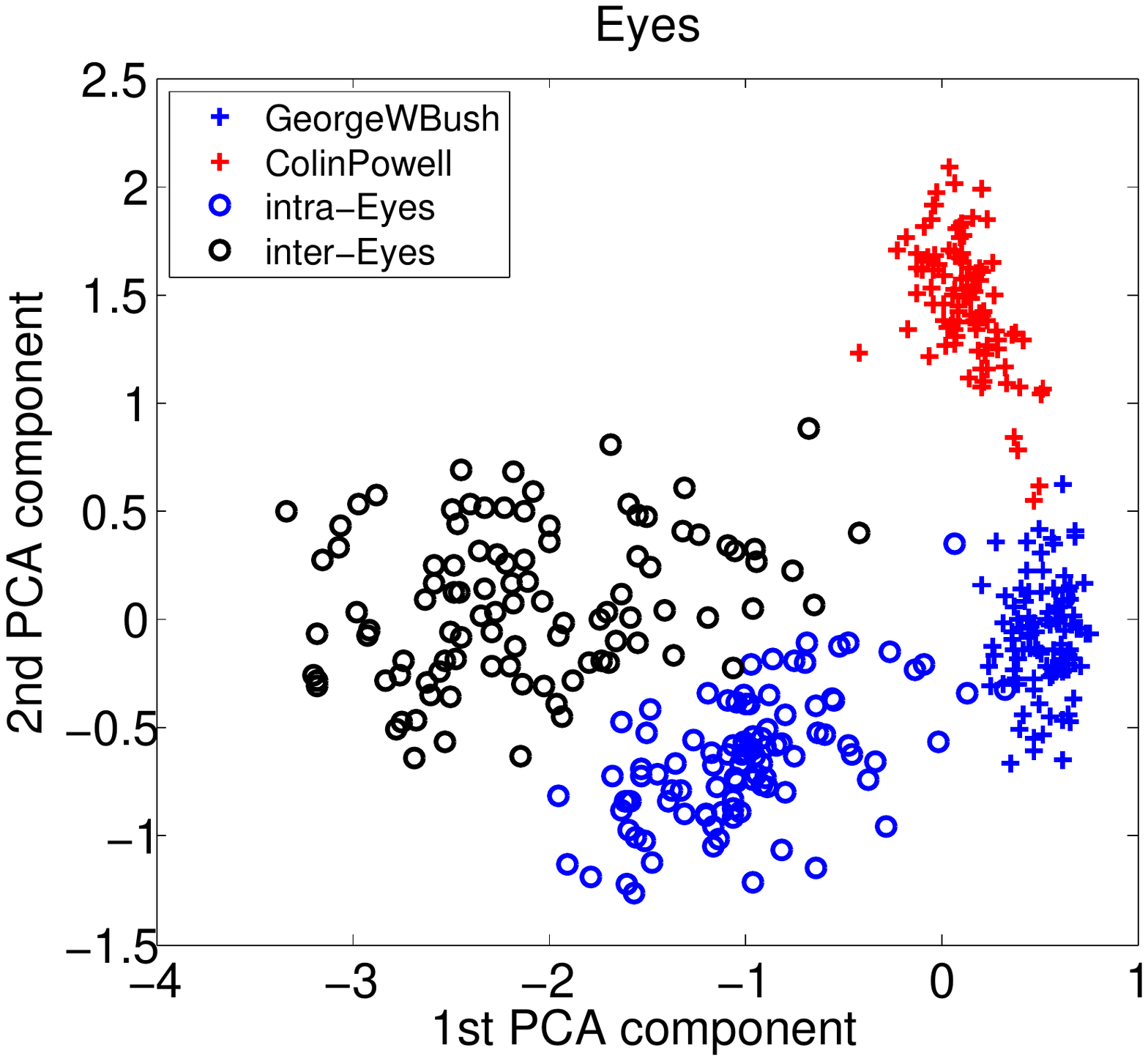}
}
\subfloat[]{
\includegraphics[trim=60 180 60 230,width=0.32\textwidth]{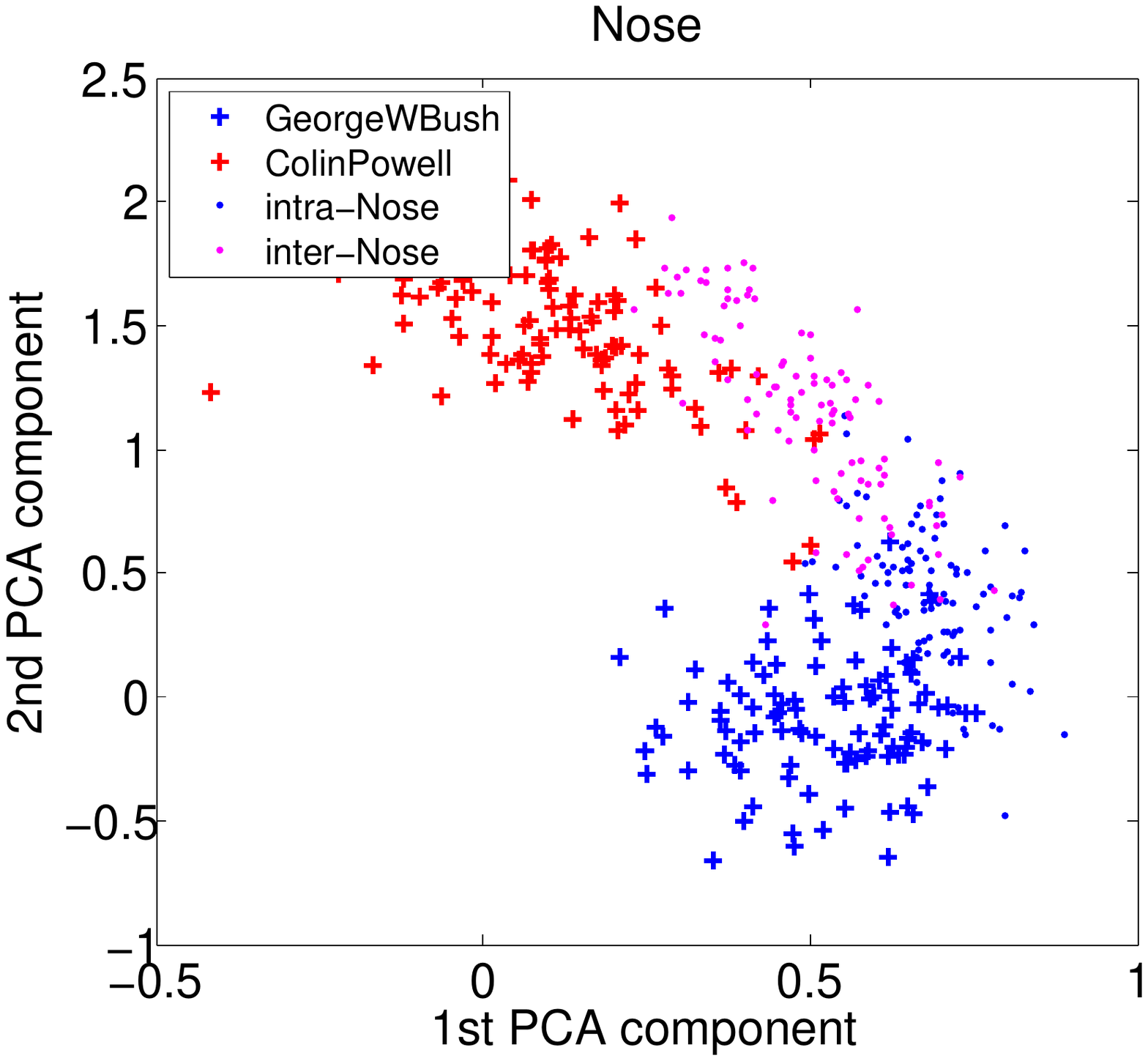}
}
\subfloat[]{
\includegraphics[trim=60 180 60 230,width=0.32\textwidth]{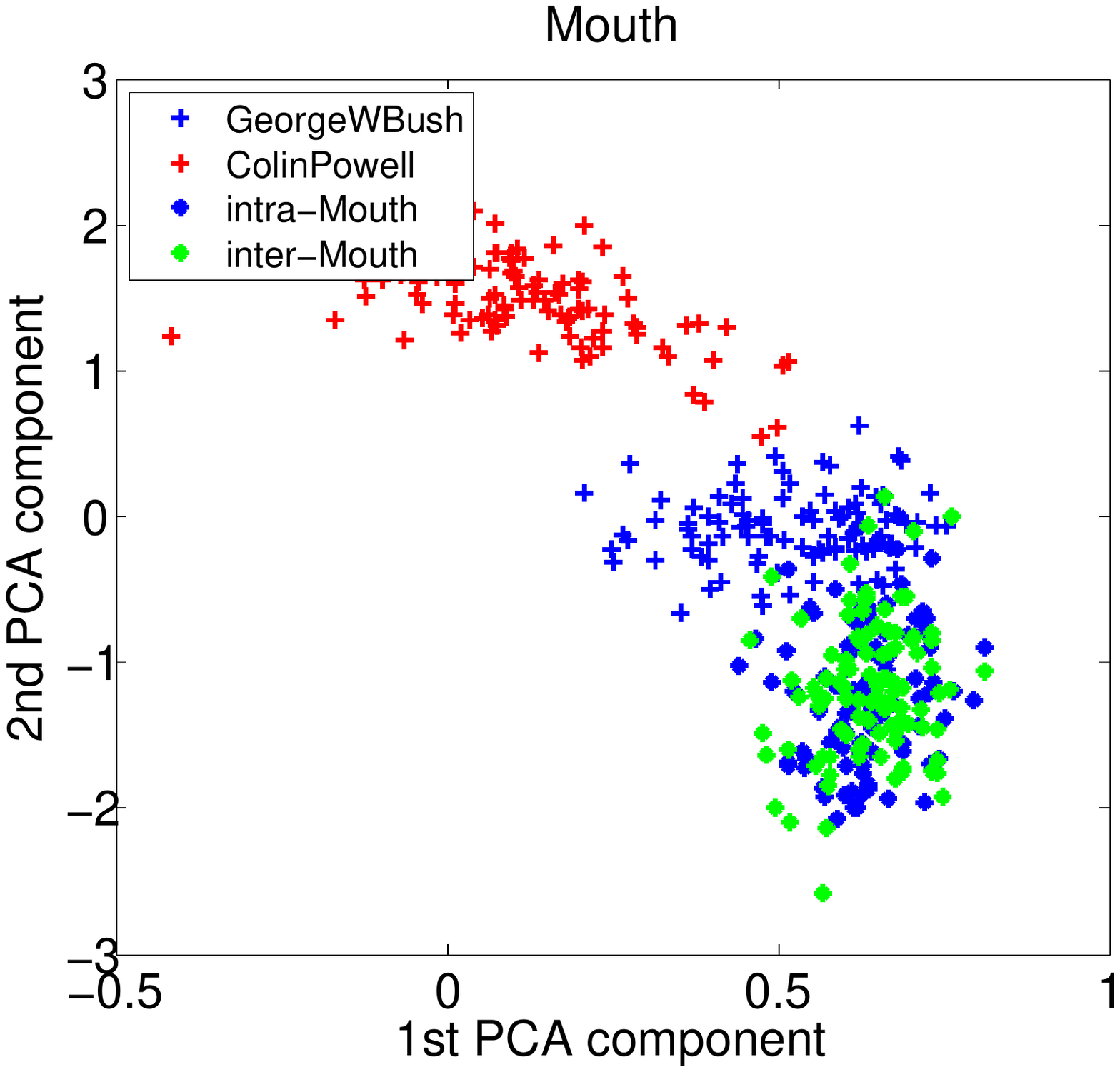}
}
\end{center}
\begin{center}
\subfloat[]{
\includegraphics[trim=10 180 70 180,width=0.45\textwidth]{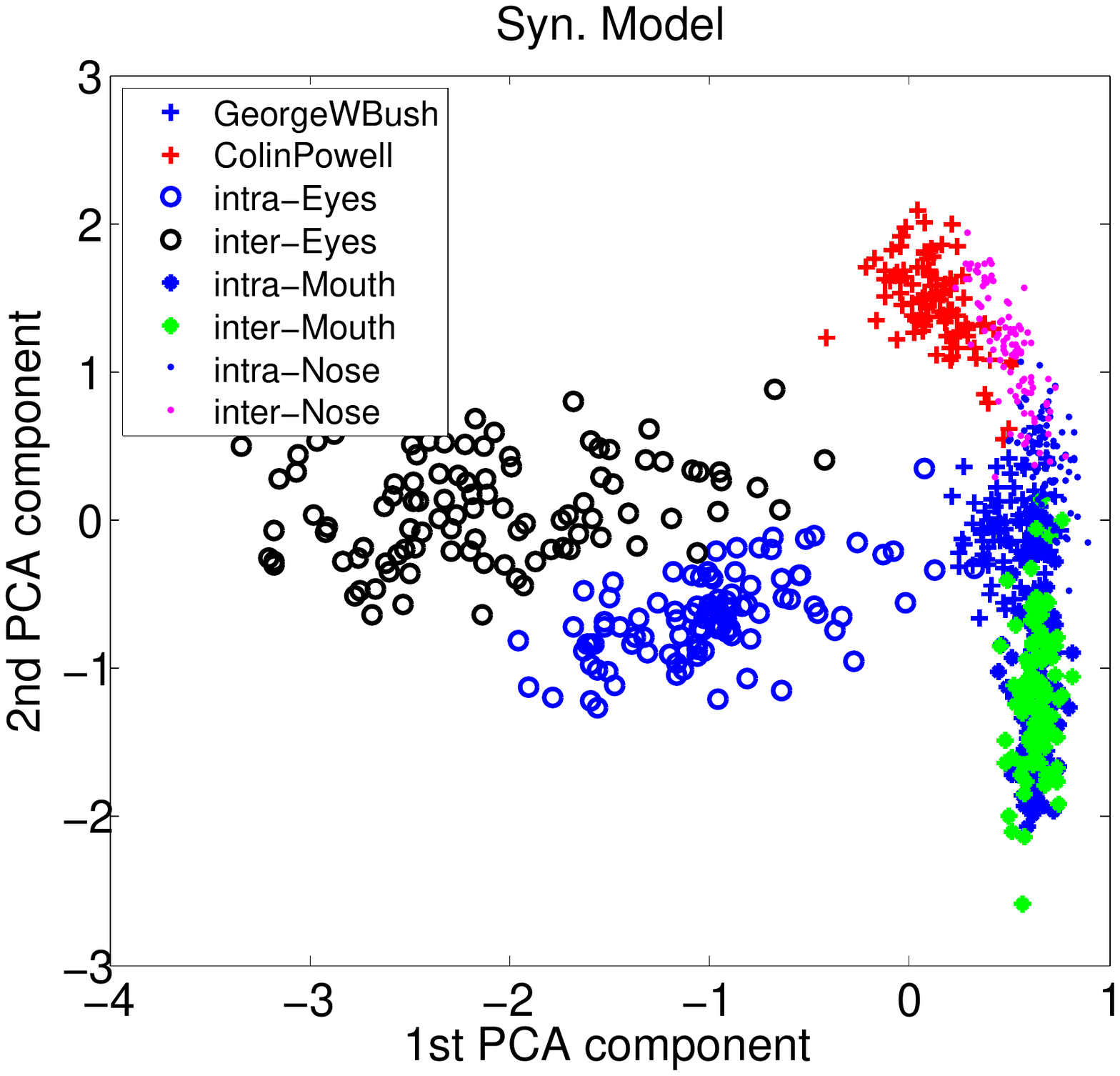}
}
\subfloat[]{
\includegraphics[trim=10 180 70 180,width=0.45\textwidth]{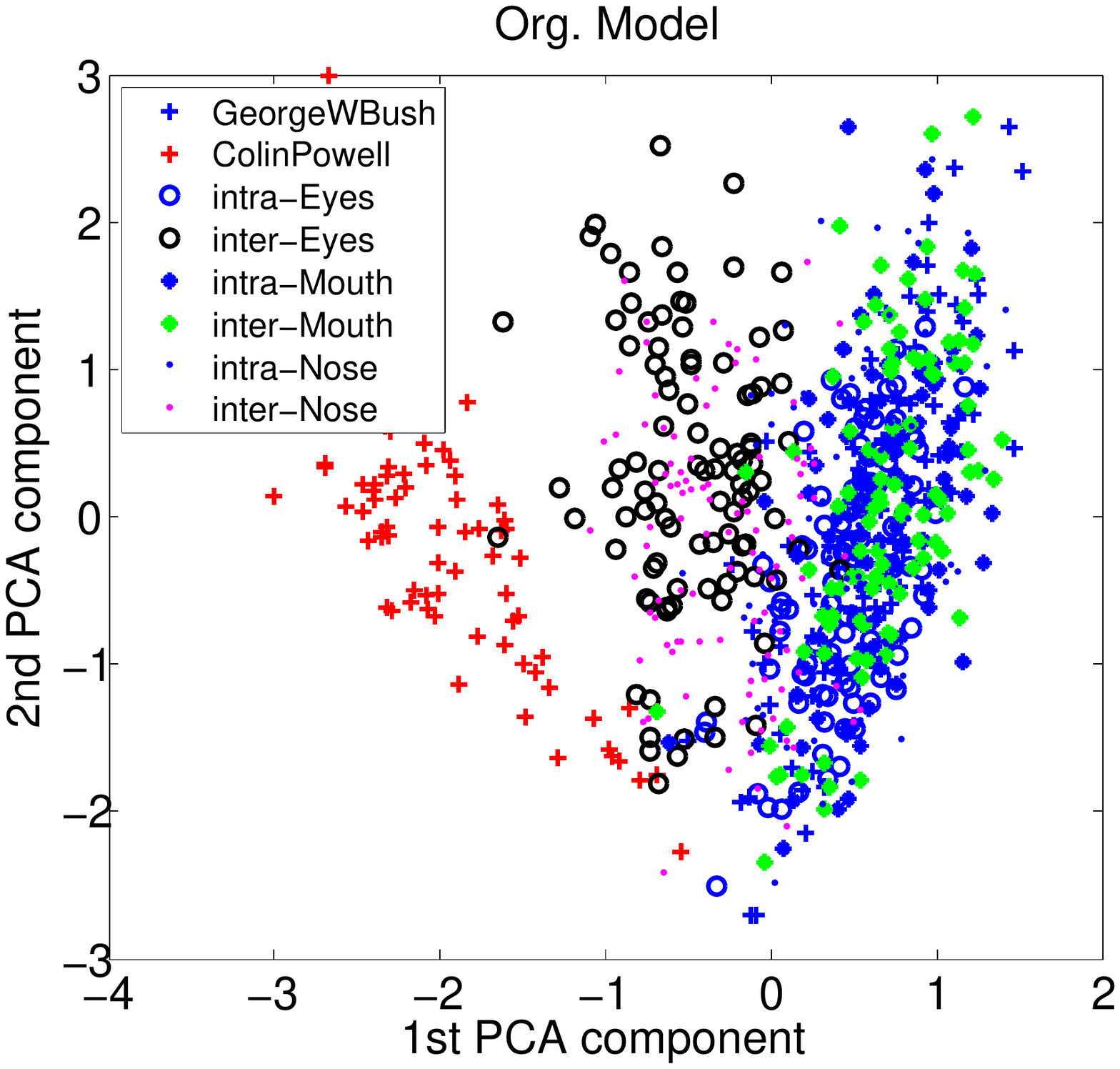}
}
\end{center}
\caption{Face distributions. `GeorgeWBush' and `ColinPowell' denote  the original images from these 2 subjects. 
`intra-X' and `inter-X' denote the synthetic images with component X replaced by another example from Bush himself and from Powell respectively, as in Fig.~\ref{fig:vissyn}. 
Features of (a)-(d) are extracted using a CNN trained using original and synthetic data, while (e) using original data only. One colour represents one (real or synthetic) subject. 
}
\label{fig:fd}
\end{figure*}

\subsubsection{Analysis of CNN model and synthetic data} 
We here analyse the trained model by visualising the synthetic images in feature space. 
We choose the images of two subjects (George W.\ Bush and Colin Powell), which have the largest number of images in LFW, 
and the synthetic images derived from the two subjects for analysis. 
To analyse the effects of replacing different facial components (eyes, nose, mouth), 
we only replace one patches of a particular facial component  of Bush with ones from himself (Intra-Synthesis) or from Powell (Inter-Synthesis), as shown in Fig. ~\ref{fig:vissyn}. 
For each case, 100 images are synthesised. Therefore, there are 8 groups of images: 2 groups of original images (Bush and Powell), 
3  Intra-Synthesis (one of three components is replaced by images of Bush) and 3  Inter-Synthesis (by images of Powell). 
These images are fed into one CNN-L to extract features, which are then projected to a PCA space. The first two PCA components of each feature are shown in Fig. ~\ref{fig:fd}. 

Fig.~\ref{fig:fd}(a)-(c) show the face distributions if one particular facial component is replaced. 
In  Fig.~\ref{fig:fd}(a), 3 identities, `GeorgeWBush+intra-Eyes', `ColinPowell' and `inter-Eyes' are well separated. 
It means that the identity information is kept if Bush's eyes are replaced by those from himself. 
In contrast, a new identity space is generated by the synthetic images that replace Bush's eyes with Powell's. 
The same conclusion can be drawn from nose in Fig.~\ref{fig:fd}(b). 
In Fig.~\ref{fig:fd}(c), however, `intra-Mouth' and `inter-Mouth' are not well separated. 
It means that the mouth component is not very discriminative between people. 
Fig.~\ref{fig:fd}(d) and (e) contrast the results when training with original and synthetic (Fig.~\ref{fig:fd}(d)) versus original data only (Fig.~\ref{fig:fd}(e)). Fig.~\ref{fig:fd}(d) is relatively more discriminative for identity, particularly the synthetic identities. Thus we can see that training with synthetic data is important to interpolate the identity space, and thus achieve good results for unseen identities -- as required at testing time.

\subsubsection{Impact of synthetic data} 
Table~\ref{tab:ISD} analyzes the importance of using the synthetic data.
First,  CNN-S trained using synthetic data (`Intra-Syn' and `Inter-Syn') outperforms greatly the model trained using just the original LFW images,
showing the importance of data synthesis.
Second, `Inter-Syn'  works slightly better than `Intra-Syn', since `Inter-Syn' can capture richer facial variations.
Third, combining `Inter-Syn' and `Intra-Syn' works better than either of them because they capture complementary variations.
Fourth, averaging the features of 32 `Self-Syn' (`32-Avg'  in Table ~\ref{tab:ISD} and defined in Section~\ref{facialdatasynt} ) images
works consistently better than that of one single test image (`single' in Table~\ref{tab:ISD} ).
Fifth, CNN-L works consistently better than CNN-S using either original LFW or synthetic images because deeper architecture has stronger generalization capacity.
Finally, metric learning further enhances the face recognition performance.


\begin{table*}[]
\centering
\caption{Comparison of synthetic data methods  on LFW}
\label{tab:ISD}
\begin{tabular}{|c| c| c|  c |c|}
\hline
  Architecture     	&\specialcell{Metric\\ learning}	&   Training data & single (\%)        & 32-Avg (\%)        \\ \hline \hline
\multirow{4}{*}{CNN-S} 	&\multirow{4}{*}{-}	&Original                        	&$ 78.97\pm0.78$ & -             \\ \cline{3-5}
			&	&Intra-Syn+Original              	&$ 83.03\pm0.56$ & $ 83.93\pm0.49 $\\ \cline{3-5}
			&	&Inter-Syn+Original         	&$ 83.18\pm0.74$ & $84.35\pm0.65$ \\ \cline{3-5}
			&	&Intra-Syn+Inter-Syn+Original & $85.61\pm0.71$ & $ 86.98\pm0.57$ \\ \hline \hline 
\multirow{6}{*}{CNN-L}  &   -   &Original                  	& $85.03\pm0.98 $ &     -          \\ \cline{2-5}
			&JB~\cite{JB}&Original         	& $87.03\pm0.69 $  		&     -          \\ \cline{2-5}
			&  -	&Intra-Syn+Inter-Syn+Original & $94.88\pm 0.66 $ & $95.13\pm0.53$ \\ \cline{2-5}			
			&\textbf{JB}~\cite{JB}	&\textbf{Intra-Syn+Inter-Syn+Original}  & $\textbf{95.32}\pm\textbf{0.38} $& $\textbf{95.77}\pm\textbf{0.38}$ \\ \cline{2-5}
			&  -	& {Intra-Syn+Inter-Syn+Original (blending) }   	& {$94.27  \pm 0.65 $}&  $ {94.46\pm 0.51} $  \\ \cline{2-5}
			&JB~\cite{JB}	&{Intra-Syn+Inter-Syn+Original (blending)} &{ $94.61  \pm 0.35 $} & {$95.05 \pm 0.34$ } \\  \hline 
\end{tabular}
\end{table*}

\subsubsection{Impact of Image Blending}  
\begin{figure}
\begin{center}
\includegraphics[trim = 50mm 45mm 40mm 25mm, clip, width=0.9 \linewidth]{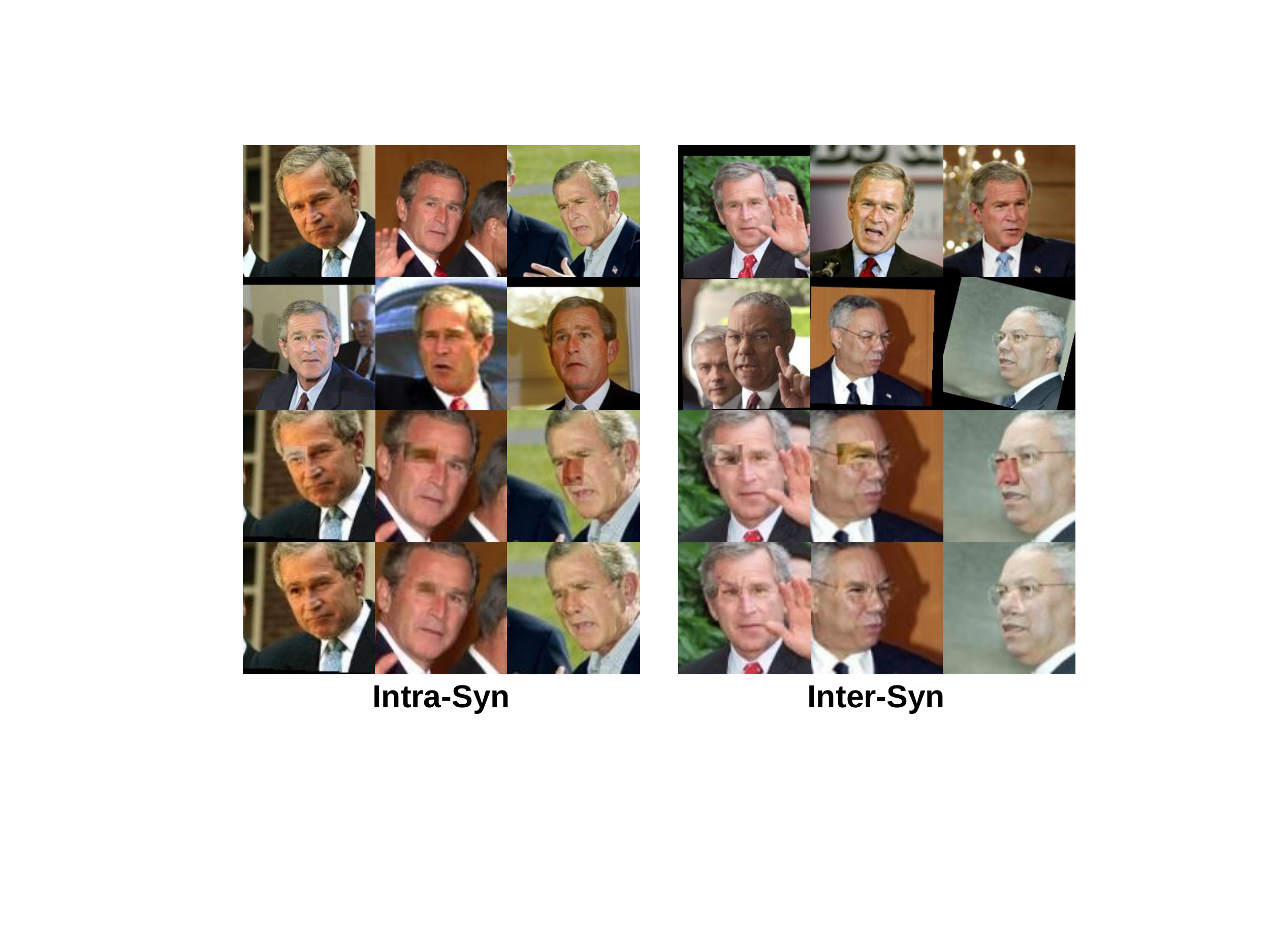}
\end{center}
\caption{Comparison of synthesis with/without blending. Row 1-2: Input image pairs, Row 3: Our synthesis with `hard boundaries', Row 4: Poisson blending} 
\label{fig:poisson}
\end{figure}

In Section \ref{sec:SDE}, we speculated the existence of artifacts (`hard boundaries') can improve the robustness of the model. 
We now experimentally investigate this by reducing such `hard  boundaries' using image blending. 
In particular, we implemented Poisson image editing ~\cite{perez2003poisson}  to smooth these boundaries. 
In Fig.~\ref{fig:poisson}, we show some results of Poisson image blending. 
From Fig.~\ref{fig:poisson}, as expected, Poisson blending does make the boundaries much smoother compared with our synthesis method. 
In Table ~\ref{tab:ISD}, we compare the results with and without Poisson image editing. 
We can see that recognition accuracy based on training data synthesised with `hard' boundaries is, somewhat, higher than with Poisson blending.  
This supports the idea that `hard' boundaries provide a source of noise that is beneficial in making  network training more robust detailed in Section~\ref{sec:SDE}.

\subsubsection{Comparison with the state-of-the-art}
Table \ref{tab:onLFW} compares our method with  state-of-the-art methods.
All methods listed in Table \ref{tab:onLFW} except ours use hand-crafted features,
again underlining the difficulty of training deep CNNs with small data.
The best deep learning solution ~\cite{sun13iccv} recorded in official benchmark achieves 91.75\%, and ours is 4\% better.
In addition, most of state-of-the-art solutions rely on extremely high dimensional feature vectors derived from densely  sampled local features on the face image.
In contrast, we just use a 320-dimensional feature vector, which is much more compact.

\begin{table}
\centering
\caption{Comparison with state-of-the-art methods  on LFW}
\begin{tabular}{|l|c|}
\hline
Methods                        & Accuracy  (\%)                \\ \hline \hline
Fisher vector faces~\cite{FVF} & $93.03 \pm 1.05 $         \\ \hline
HPEN    ~\cite{zhu2015high}		& $95.25 \pm 0.36$		\\ \hline
MDML-DCPs ~\cite{ding2014multi}             & $95.58 \pm 0.34 $          \\ \hline
\textbf{The proposed}    	& $\textbf{95.77} \pm \textbf{0.38} $\\ \hline
\end{tabular}
\label{tab:onLFW}
\end{table}

\subsubsection{Non-CNN methods using synthetic data} 
Above we demonstrated the effectiveness of  synthetic data to train CNNs. 
We now consider its usefulness to improving methods based on traditional hand-crafted features. 
One typical hand-crafted feature used for unconstrained face recognition problem is high dimensional LBP (HD-LBP)~\cite{HimF}. 
We extract the HD-LBP feature using the open source code ~\cite{hdlbptw}. 
From Table~\ref{tab:hdlbp}, we see that HD-LBP with JB metric learning trained using original LFW images works much better that without JB (89.02\% vs 84.13\%), showing the expected effectiveness of metric learning. More interestingly, we see that training JB using both original  and synthetic images outperforms that using original images only, 91.03\% vs 89.02\%. This shows that our synthetic data approach is also useful in combination with conventional hand-crafted features.

\begin{table}
\centering
\begin{threeparttable}
\caption{Hand-crafted features  on LFW}
\label{tab:hdlbp}
\begin{tabular}{|c|c|}
\hline
Methods                        	& Accuracy  (\%)                \\ \hline \hline
HD-LBP			       	& $84.13 \pm 1.76$          \\ \hline
HD-LBP+JB (original) 	& $ 89.02 \pm 1.11$          \\ \hline
HD-LBP+JB (original + synthetic)& $91.03 \pm 1.06 $          \\ \hline
\end{tabular}
\end{threeparttable}
\end{table}

\subsubsection{Impact of synthetic images on larger training sets} 
\begin{figure} 
\begin{center}
\includegraphics[trim = 30mm 85mm 20mm 85mm, clip, width=0.85 \linewidth]{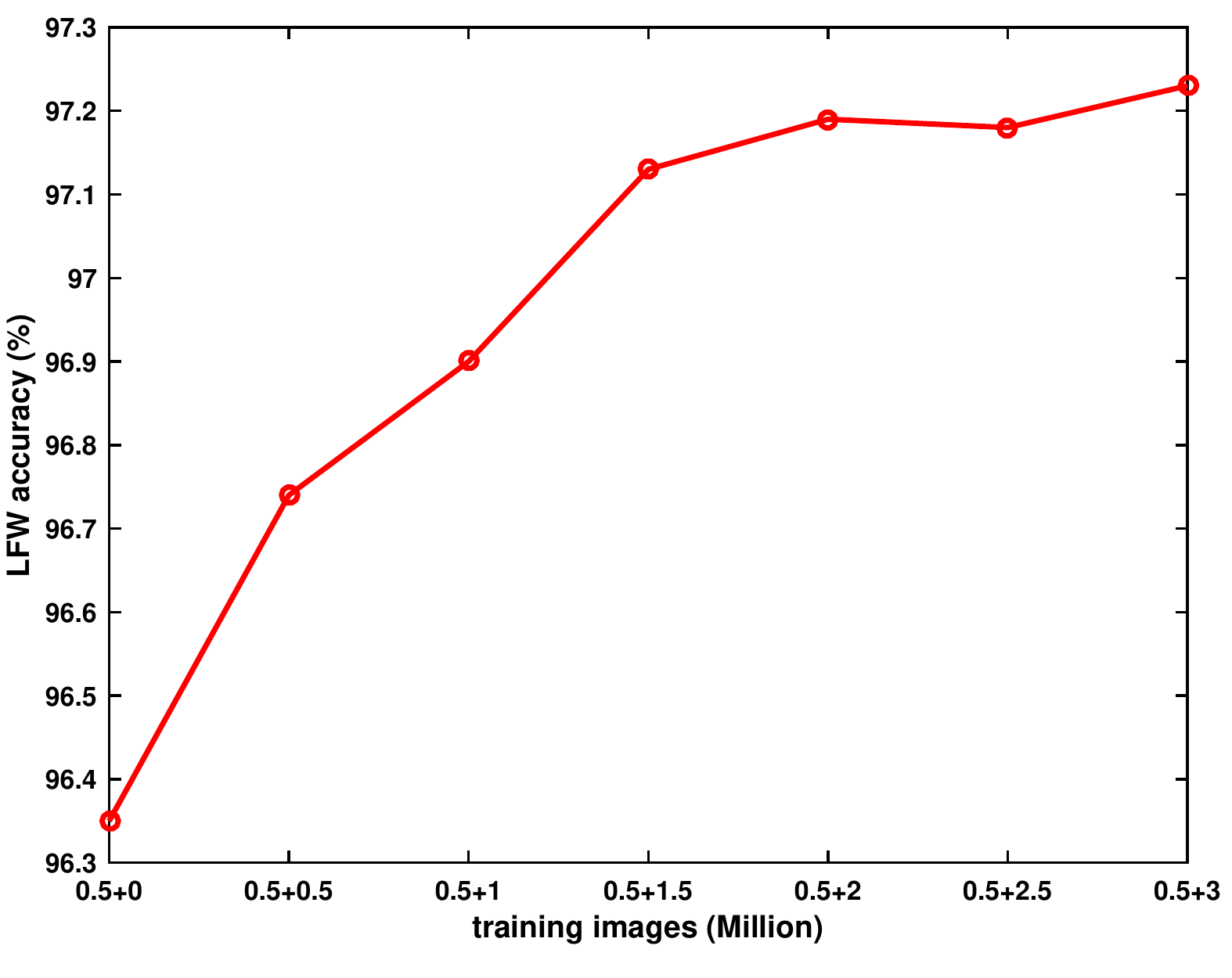}
\end{center}
\caption{Recognition rate (\%) on LFW using the original  CASIA WEBFACE training data (0.5M) and synthetic data (0-3M).} 
\label{fig:bigsyn}
\end{figure}

Our original motivation was to learn effective face representations from small datasets. 
We demonstrated the effectiveness of generating synthetic data to expand a small training set (LFW, 5K identities and 10K images) in Table~\ref{tab:ISD}. 
Recently, some bigger training sets of face images in the wild have been released, such as CASIA WEBFACE~\cite{WEBFACE} (10K identities and 0.5M images). 
We conduct  experiments using the latter dataset to assess whether our data synthesis strategy is also useful for such larger datasets.
As before, we keep the ratio 2:1 of `Inter-Syn' and `Intra-Syn' synthetic images. 
To investigate the effects of the size of synthetic data, we generate  six sets of synthetic images:  \{0.5M, 1M, 1.5M, 2M, 2.5M, 3M\} images. 
We trained the CNN-L network using the original CASIA WEBFACE images, plus a variable amount of synthetic data. 
From Fig~\ref{fig:bigsyn}, we can see that the  recognition rate on LFW  increases with the amount of synthetic data. 
This demonstrates that our data synthesis strategy is still very effective even with relatively large datasets. 

\subsection{NIR-VIS face recognition}
\label{sec:ExpNirVis}
\subsubsection{Database and protocol} The largest face database across NIR and VIS spectrum so far is the CASIA NIR-VIS 2.0 face database~\cite{NIRVIS}.
It contains 17,580 images of 725 subjects which exhibit intra-personal variations such as pose and expression.
This database includes two views: view 1 for  parameter tuning and view 2 including 10 folds for performance evaluation.
During test, the gallery and probe images are VIS and NIR images respectively,  simulating the scenario of face recognition in the dark environment.
The rank 1 identification rate including the mean accuracy and standard deviation of 10 folds are reported.

Because the images of CASIA NIR-VIS2.0 are from two modalities (NIR and VIS),
we applied `cross-modality synthesis' to synthesize new images.
The size of synthesized data is detailed in Table ~\ref{tab:SDG}.

\begin{table}
\centering
\caption{Synthetic data using CASIA NIR-VIS2.0 database}
\label{tab:SDG}
\begin{tabular}{|c|c|c|c|c|}
\hline
	  &   			& IDs & Images & Images/ID \\ \hline
\multirow{3}{*}{Synthetic}	&Intra-Syn & 357       & 90K   &       250     \\ \cline{2-5}
			  	&Inter-Syn & 1K       & 150K     &      150      \\ \cline{2-5}
			   	&Total     & 1.4K      & 240K   &       170     \\ \hline
\multicolumn{2}{|c|}{Original}	& 357    &   8.5K     & 23 \\ \hline
\end{tabular}
\end{table}

\subsubsection{Illumination normalization and feature extraction}

\begin{figure} 
\begin{center}
\includegraphics[ clip, width=0.8 \linewidth]{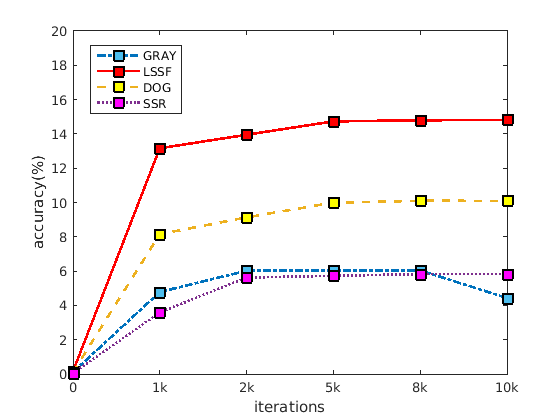}\\
\includegraphics[ width=0.8 \linewidth]{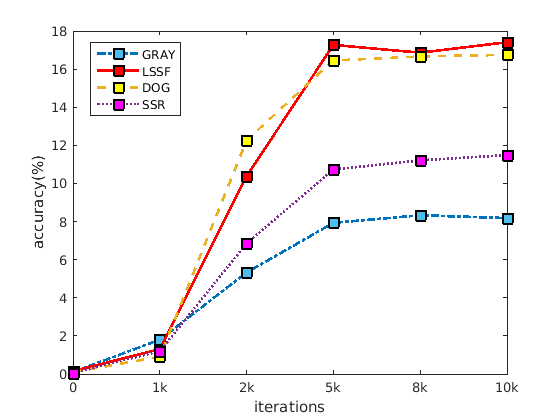}
\end{center}
\caption{Effect of illumination normalization methods using the CNN-S (top) and CNN-L networks.}
\label{fig:cinm}
\end{figure}

Illumination Normalization (IN) methods are usually used to narrow the gap between NIR and VIS images.
To investigate the impact of IN, we preprocessed images using three popular IN methods:
illumination normalization based on large-and small-scale features (LSSF)~\cite{LSSF},  Diffence-of-Gaussian filtering-based normalization (DOG), and single-scale retinex (SSR)~\cite{SSR}.
We train CNN-S and CNN-L using  illumination normalized  and non-normalized images.
For simplicity, only the images from CASIA NIR-VIS2.0 excluding synthetic ones are used.
Fig.~\ref{fig:cinm} shows the face recognition rates at different training iterations using different input images for the CNN-S and CNN-L networks.
The results show the effectiveness of IN, and LSSF achieves the best performance due to its strong capacity of removing illumination effects without affecting identity information.
As for the LFW experiments in Section~\ref{sec:ExpFRW}, CNN-L works better than CNN-S.

We also extracted LBP features from the LSSF-normalized images, and achieve $12.48\pm3.1$ in comparison with $ 17.41 \pm 3.76 $ obtained with features from the CNN-L network.
Showing again the superior performance of CNN learned features.

\begin{table}
\centering
\caption{Evaluation of the impact of synthetic data}
\label{tab:COTPWWSD}
\begin{tabular}{c|c|c|c|}
\cline{2-4}
\multirow{2}{*}{}                                     & \multicolumn{2}{c|}{Training Data}  & \multirow{2}{*}{Accuracy(\%)} \\ \cline{2-3}
                                                      & \specialcell{CASIA \\ NIR-VIS2.0 }  & LFW             &                               \\ \hline
\multicolumn{1}{|c|}{Baseline}                        			 & Raw              & -               & $ 17.41 \pm 3.76 $             \\ \hline
\multicolumn{1}{|c|}{\multirow{4}{*}{\specialcell{Synthetic\\ Data}}}    & Raw+Syn          & -               & $ 34.13 \pm 2.13 $              \\ \cline{2-4}
\multicolumn{1}{|c|}{}                                			 &  -               & Raw             & $38.45 \pm 2.08  $              \\    \cline{2-4}
\multicolumn{1}{|c|}{}                                			 & -                & Raw+Syn         & $66.37 \pm 1.45 $  \\ \cline{2-4}
\multicolumn{1}{|c|}{}                                			 & Raw+Syn          & Raw+Syn         & $\textbf{68.97} \pm \textbf{1.24} $  \\ \hline

\end{tabular}
\end{table}

%

\subsubsection{Effects of synthetic data}
In practice, we find two problems with the synthetic data generated from the CASIA NIR-VIS2.0 dataset:
(1) It cannot capture enough facial variations because it only has 357 subjects as shown in Table~\ref{tab:SDG}.
(2) There are much fewer VIS images than NIR ones.
To solve these two problems,  we also use the synthetic data generated from LFW images defined in Table~\ref{tab:SDGLFW}.

Table ~\ref{tab:COTPWWSD} compares the results achieved by these two sources of synthetic data.
First,  the accuracy achieved by using the synthetic data generated from CASIA NIR-VIS2.0 database
is  $ 34.13 \pm 2.13 $, in comparison with $ 17.41 \pm 3.76 $ without synthetic data.
The significant improvement shows the effectiveness of data synthesis.
Second, the model trained using raw and synthetic LSSF-normalized LFW images greatly
outperforms those synthetic CASIA  NIR-VIS2.0 images: $66.37 \pm 1.45 $  vs.\ $38.45 \pm 2.08  $,
although NIR images are completely unseen during training.
The reasons are 2-fold:
(1) LFW images contains more subjects which can capture more facial variations as analyzed above.
(2) LSSF can greatly reduce the gap between NIR and VIS, therefore, LSSF-normalized LFW synthetic images  can generalize well to LSSF-normalized NIR images.
To further improve the face recognition performance, we trained the network using both raw and synthetic data from both sources (LFW+CASIA NIR-VIS).
The face recognition rate is improved from $66.37 \pm 1.45 $ to $68.97 \pm 1.24$, showing the value provided by our of bigger synthetic dataset.

\begin{table}
\centering
\caption{Comparisons with the state of the art on the CASIA NIR-VIS2.0 dataset.}
\label{tab:CCVAS}
\resizebox{\columnwidth}{!}{
\begin{tabular}{|c|c|c|l|}
\hline
\multicolumn{2}{|c|}{Method}                                &            \multicolumn{2}{|c|}{Accuracy (\%)}              \\ \hline \hline
\multirow{4}{*}{CNN-L}            & \multirow{2}{*}{Training Data} & Original            & $69.11 \pm 1.21$ \\ \cline{3-4}
                                  &                                & LSSF                & $68.97 \pm 1.24$ \\ \cline{2-4}
                                  & Network Fusion                 & Original+LSSF       & $79.96 \pm 1.18$ \\ \cline{2-4}
                                  & Metric Learning                & LDA (Original+LSSF) & $\textbf{85.05} \pm \textbf{0.83}$ \\ \hline \hline
\multirow{3}{*}{State-of-the-art} & \multicolumn{2}{c|}{C-CBFD~\cite{lulearning}}        & $56.6 \pm 2.4$ \\ 
 \cline{2-4} & \multicolumn{2}{c|}{Dictionary Learning~\cite{juefei2015nir}} & $78.46 \pm 1.67$ \\ 
\cline{2-4}  & \multicolumn{2}{c|}{C-CBFD + LDA~\cite{lulearning}}        & $81.8 \pm 2.3$ \\ 
\cline{2-4}  & \multicolumn{2}{c|}{CNN + LDML~\cite{saxena16eccv}}        & $ 85.9 \pm 0.9$ \\ 
\cline{2-4}  & \multicolumn{2}{c|}{Gabor + RBM~\cite{yi15shared}}        & $\bf 86.2 \pm 1.0$ \\ 
\hline
\end{tabular}
}
\end{table}

\subsubsection{Comparison with the state-of-the-art}
The CNN-L models in Table~\ref{tab:CCVAS} are  all trained using synthetic LFW data.
First, LSSF-normalized and Original LFW synthetic data achieve very comparable performance: 68.97\% vs.\ 69.11\%.
However, the fusion (averaging) of these two features can significantly improve the face recognition rates.
It shows the fusion can keep the discriminative facial information but remove the illumination effects.
Second, not surprisingly, metric learning can further improve the performance.
The metric learning method used here is LDA, which is the most widely used one for face identification.
Finally, Table~\ref{tab:CCVAS} compares the proposed method against the state-of-the-art solutions \cite{lulearning, juefei2015nir}.
\cite{lulearning} uses a designed descriptor that performs better in this dataset compared with other generic hand-crafted features,
and LDA can further improve the accuracy.
Our method significantly outperforms \cite{lulearning} when metric learning is not used (79.96\% vs.\ 56.6\%), and maintains superior performance when metric learning is used (85.05\% vs.\ 81.8\%).
\cite{juefei2015nir} tries to solve the domain shift between two data sources (NIR and VIS) by a cross-modal metric learning:
it assumes that a pair of NIR and VIS images shares the same sparse representation under two jointly learned dictionaries.
Our method improves over \cite{juefei2015nir} with a 7\% margin without such an extra step of dictionary learning.
Concurrently to our work, Saxena \& Verbeek \cite{saxena16eccv} obtained a comparable performance of 85.9\% using a cross-modal version of LDML metric learning \cite{guillaumin09iccv1}, albeit using  CNN features learned from the 500,000 face images of the CASIA WEBFACE dataset.
Finally, Yi et al.~\cite{yi15shared} obtained the best performance of 86.1\% using an approach that extracts 40 dim.\ Gabor features at 176 local face regions, and uses these to train 176 different restricted Boltzmann machines (RBMs) specialized to model the modality shift at each face region. Note that unlike our approach based on feed-forward CNNs, their approach requires Gibbs sampling at test-time to infer the face representations.


\section{Conclusions and Future Work}
Recently,  convolutional neural networks have attracted
a  lot  of  attention  in  the  field  of  face  recognition. However, deep learning methods heavily depend on big training data, which is not always available.
To solve this problem in the field of face recognition, we propose a new face synthesis method which swaps the facial components of different face images to generate a new face.
With this technique, we achieve state-of-the-art face recognition performance on LFW and CASIA NIR-VIS2.0 face databases.

In the future, we will apply this technique to more applications of face analysis. 
For example, the proposed data synthesis method can easily be used in training CNN-based face detection, facial attribute recognition, etc.
More generally, the method applies to any objects which are well structured.  For example, the human body is well structured and human images can be synthesised using this method.  The synthetic images can be used to train deep models for pose estimation, pedestrian detection, and person re-identifiation.

\section*{Acknowledgment}
 This work is supported by European Union’s Horizon 2020 research and innovation program under grant No 640891, 
 the Science and Technology Plan Project of Hunan Province (2016TP1020), 
the Natural Science Foundation of China (No. 61502152), and the French research agency contracts ANR-16-CE23-0006 and ANR-11-LABX-0025-01. 
We gratefully acknowledge NVIDIA for the donation of the GPUs for this research.


\begin{thebibliography}{10}\itemsep=-1pt

\bibitem{LBP}
T.~Ahonen, A.~Hadid, and M.~Pietik{\"a}inen.
\newblock Face recognition with local binary patterns.
\newblock In {\em Computer vision-eccv 2004}, pages 469--481. Springer, 2004.

\bibitem{LPQ}
T.~Ahonen, E.~Rahtu, V.~Ojansivu, and J.~Heikkila.
\newblock Recognition of blurred faces using local phase quantization.
\newblock In {\em International Conference on Pattern Recognition}, 2008.

\bibitem{belhumeur97pami}
P.~Belhumeur, J.~Hespanha, and D.~Kriegman.
\newblock Eigenfaces vs.\ {F}isherfaces: Recognition using class specific
  linear projection.
\newblock {\em PAMI}, 19(7):711--720, 1997.

\bibitem{hdlbptw}
B.-C. Chen, C.-S. Chen, and W.~H. Hsu.
\newblock Review and implementation of high-dimensional local binary patterns
  and its application to face recognition.
\newblock Technical Report TR-IIS-14-003, Institute of Information Science,
  Academia Sinica, 2014.

\bibitem{JB}
D.~Chen, X.~Cao, L.~Wang, F.~Wen, and J.~Sun.
\newblock Bayesian face revisited: A joint formulation.
\newblock In {\em ECCV}. 2012.

\bibitem{HimF}
D.~Chen, X.~Cao, F.~Wen, and J.~Sun.
\newblock Blessing of dimensionality: High-dimensional feature and its
  efficient compression for face verification.
\newblock In {\em CVPR}, 2013.

\bibitem{decoste02ml}
D.~Decoste and B.~Sch\"olkopf.
\newblock Training invariant support vector machines.
\newblock {\em Machine Learning}, 46:161--190, 2002.

\bibitem{ding2014multi}
C.~Ding, J.~Choi, D.~Tao, and L.~S. Davis.
\newblock Multi-directional multi-level dual-cross patterns for robust face
  recognition.
\newblock {\em arXiv preprint arXiv:1401.5311}, 2014.

\bibitem{feng15tip}
Z.~Feng, G.~Hu, J.~Kittler, W.~Christmas, and X.~Wu.
\newblock Cascaded collaborative regression for robust facial landmark
  detection trained using a mixture of synthetic and real images with dynamic
  weighting.
\newblock {\em IEEE Transactions on Image Processing}, 24(11):3425--3440, 2015.

\bibitem{girshick14cvpr}
R.~Girshick, J.~Donahue, T.~Darrell, and J.~Malik.
\newblock Rich feature hierarchies for accurate object detection and semantic
  segmentation.
\newblock In {\em CVPR}, 2014.

\bibitem{guillaumin09iccv1}
M.~Guillaumin, J.~Verbeek, and C.~Schmid.
\newblock Is that you? {M}etric learning approaches for face identification.
\newblock In {\em ICCV}, 2009.

\bibitem{hassner2014effective}
T.~Hassner, S.~Harel, E.~Paz, and R.~Enbar.
\newblock Effective face frontalization in unconstrained images.
\newblock {\em arXiv preprint arXiv:1411.7964}, 2014.

\bibitem{he2015deep}
K.~He, X.~Zhang, S.~Ren, and J.~Sun.
\newblock Deep residual learning for image recognition.
\newblock {\em arXiv preprint arXiv:1512.03385}, 2015.

\bibitem{he05pami}
X.~He, S.~Yan, Y.~Hu, P.~Niyogi, and H.-J. Zhang.
\newblock Face recognition using {L}aplacianfaces.
\newblock {\em PAMI}, 27(3):328--340, 2005.

\bibitem{huattribute}
G.~Hu, Y.~Hua, Y.~Yuan, Z.~Zhang, Z.~Lu, S.~S. Mukherjee, T.~M. Hospedales,
  N.~M. Robertson, and Y.~Yang.
\newblock Attribute-enhanced face recognition with neural tensor fusion
  networks.
\newblock 2017.

\bibitem{hu2016face}
G.~Hu, F.~Yan, C.-H. Chan, W.~Deng, W.~Christmas, J.~Kittler, and N.~M.
  Robertson.
\newblock Face recognition using a unified 3d morphable model.
\newblock In {\em European Conference on Computer Vision}, pages 73--89.
  Springer, 2016.

\bibitem{hu2017efficient}
G.~Hu, F.~Yan, J.~Kittler, W.~Christmas, C.~H. Chan, Z.~Feng, and P.~Huber.
\newblock Efficient 3d morphable face model fitting.
\newblock {\em Pattern Recognition}, 67:366--379, 2017.

\bibitem{MyDLeval}
G.~Hu, Y.~Yang, D.~Yi, J.~Kittler, W.~J. Christmas, S.~Z. Li, and T.~M.
  Hospedales.
\newblock When face recognition meets with deep learning: an evaluation of
  convolutional neural networks for face recognition.
\newblock {\em CoRR}, abs/1504.02351, 2015.

\bibitem{huang07lfw}
G.~Huang, M.~Ramesh, T.~Berg, and E.~Learned-Miller.
\newblock Labeled faces in the wild: a database for studying face recognition
  in unconstrained environments.
\newblock Technical Report 07-49, University of Massachusetts, Amherst, 2007.

\bibitem{jaderberg2014syn}
M.~Jaderberg, K.~Simonyan, A.~Vedaldi, and A.~Zisserman.
\newblock Synthetic data and artificial neural networks for natural scene text
  recognition.
\newblock {\em arXiv preprint arXiv:1406.2227}, 2014.

\bibitem{jegou10cvpr}
H.~J{\'e}gou, M.~Douze, C.~Schmid, and P.~P{\'e}rez.
\newblock Aggregating local descriptors into a compact image representation.
\newblock In {\em CVPR}, 2010.

\bibitem{SSR}
D.~J. Jobson, Z.-U. Rahman, and G.~A. Woodell.
\newblock Properties and performance of a center/surround retinex.
\newblock {\em IEEE Trans. Image Processing}, 6(3):451--462, 1997.

\bibitem{juefei2015nir}
F.~Juefei-Xu, D.~Pal, and M.~Savvides.
\newblock Nir-vis heterogeneous face recognition via cross-spectral joint
  dictionary learning and reconstruction.
\newblock In {\em Proceedings of the IEEE Conference on Computer Vision and
  Pattern Recognition Workshops}, pages 141--150, 2015.

\bibitem{kong05cviu}
S.~Kong, J.~Heo, B.~Abidi, J.~Paik, and M.~Abidi.
\newblock Recent advances in visual and infrared face recognition -- a review.
\newblock {\em CVIU}, 97(1):103 -- 135, 2005.

\bibitem{krizhevsky12nips}
A.~Krizhevsky, I.~Sutskever, and G.~Hinton.
\newblock Imagenet classification with deep convolutional neural networks.
\newblock In {\em NIPS}, 2012.

\bibitem{NIRVIS}
S.~Z. Li, D.~Yi, Z.~Lei, and S.~Liao.
\newblock The casia nir-vis 2.0 face database.
\newblock In {\em Computer Vision and Pattern Recognition Workshops (CVPRW),
  2013 IEEE Conference on}, pages 348--353. IEEE, 2013.

\bibitem{BaiduFace}
J.~Liu, Y.~Deng, and C.~Huang.
\newblock Targeting ultimate accuracy: Face recognition via deep embedding.
\newblock {\em arXiv preprint arXiv:1506.07310}, 2015.

\bibitem{lowe04ijcv}
D.~Lowe.
\newblock Distinctive image features from scale-invariant keypoints.
\newblock {\em IJCV}, 60(2):91--110, 2004.

\bibitem{lulearning}
J.~Lu, V.~E. Liong, X.~Zhou, and J.~Zhou.
\newblock Learning compact binary face descriptor for face recognition.
\newblock {\em Pattern Analysis and Machine Intelligence, IEEE Transactions
  on}, 2015.

\bibitem{masi2016we}
I.~Masi, A.~T. Tran, J.~T. Leksut, T.~Hassner, and G.~Medioni.
\newblock Do we really need to collect millions of faces for effective face
  recognition?
\newblock {\em ECCV}, 2016.

\bibitem{megaface}
D.~Miller, E.~Brossard, S.~M. Seitz, and I.~Kemelmacher-Shlizerman.
\newblock Megaface: A million faces for recognition at scale.
\newblock {\em arXiv preprint arXiv:1505.02108}, 2015.

\bibitem{ouyang2014hfrSurvey}
S.~Ouyang, T.~Hospedales, Y.-Z. Song, and X.~Li.
\newblock A survey on heterogeneous face recognition: Sketch, infra-red, 3d and
  low-resolution.
\newblock {\em arXiv preprint arXiv:1409.5114}, 2016.

\bibitem{papon2015semantic}
J.~Papon and M.~Schoeler.
\newblock Semantic pose using deep networks trained on synthetic rgb-d.
\newblock {\em arXiv preprint arXiv:1508.00835}, 2015.

\bibitem{parkhi15bmvc}
O.~Parkhi, A.~Vedaldi, and A.~Zisserman.
\newblock Deep face recognition.
\newblock In {\em BMVC}, 2015.

\bibitem{paulin14cvpr}
M.~Paulin, J.~Revaud, Z.~Harchaoui, F.~Perronnin, and C.~Schmid.
\newblock Transformation pursuit for image classification.
\newblock In {\em CVPR}, 2014.

\bibitem{perez2003poisson}
P.~P{\'e}rez, M.~Gangnet, and A.~Blake.
\newblock Poisson image editing.
\newblock In {\em ACM Transactions on Graphics}, 2003.

\bibitem{pinheiro15cvpr}
P.~Pinheiro and R.~Collobert.
\newblock From image-level to pixel-level labeling with convolutional networks.
\newblock In {\em CVPR}, 2015.

\bibitem{rogez16nips}
G.~Rogez and C.~Schmid.
\newblock {MoCap}-guided data augmentation for {3D} pose estimation in the
  wild.
\newblock In {\em NIPS}, 2016.

\bibitem{rozantsev2015rd}
A.~Rozantsev, V.~Lepetit, and P.~Fua.
\newblock On rendering synthetic images for training an object detector.
\newblock {\em Computer Vision and Image Understanding}, 2015.

\bibitem{rozantsev15cviu}
A.~Rozantsev, V.~Lepetit, and P.~Fua.
\newblock On rendering synthetic images for training an object detector.
\newblock {\em CVIU}, 137:24 -- 37, 2015.

\bibitem{sanchez13ijcv}
J.~S\'{a}nchez, F.~Perronnin, T.~Mensink, and J.~Verbeek.
\newblock Image classification with the {F}isher vector: Theory and practice.
\newblock {\em IJCV}, 105(3):222--245, 2013.

\bibitem{saxena16eccv}
S.~Saxena and J.~Verbeek.
\newblock Heterogeneous face recognition with {CNNs}.
\newblock In {\em ECCV TASK-CV Workshop}, 2016.

\bibitem{schroff15cvpr}
F.~Schroff, D.~Kalenichenko, and J.~Philbin.
\newblock Facenet: A unified embedding for face recognition and clustering.
\newblock In {\em CVPR}, 2015.

\bibitem{shotton13acm}
J.~Shotton, T.~Sharp, A.~Kipman, A.~Fitzgibbon, M.~Finocchio, A.~Blake,
  M.~Cook, and R.~Moore.
\newblock Real-time human pose recognition in parts from single depth images.
\newblock {\em Communications of the ACM}, 56(1):116--124, 2013.

\bibitem{sim02fg}
T.~Sim, S.~Baker, and M.~Bsat.
\newblock The {CMU Pose, Illumination, and Expression (PIE)} database.
\newblock In {\em IEEE International Conference on Automatic Face and Gesture
  Recognition}, 2002.

\bibitem{FVF}
K.~Simonyan, O.~M. Parkhi, A.~Vedaldi, and A.~Zisserman.
\newblock {F}isher {V}ector {F}aces in the {W}ild.
\newblock In {\em British Machine Vision Conference}, 2013.

\bibitem{simonyan14nips}
K.~Simonyan and A.~Zisserman.
\newblock Two-stream convolutional networks for action recognition in videos.
\newblock In {\em NIPS}, 2014.

\bibitem{simonyan15iclr}
K.~Simonyan and A.~Zisserman.
\newblock Very deep convolutional networks for large-scale image recognition.
\newblock In {\em ICLR}, 2015.

\bibitem{su15iccv}
H.~Su, C.~Qi, Y.~Li, and L.~Guibas.
\newblock Render for cnn: Viewpoint estimation in images using cnns trained
  with rendered 3d model views.
\newblock In {\em ICCV}, 2015.

\bibitem{sun13iccv}
C.~Sun and R.~Nevatia.
\newblock {ACTIVE:} activity concept transitions in video event classification.
\newblock In {\em ICCV}, 2013.

\bibitem{DEEPID2}
Y.~Sun, Y.~Chen, X.~Wang, and X.~Tang.
\newblock Deep learning face representation by joint
  identification-verification.
\newblock In {\em Advances in Neural Information Processing Systems}, pages
  1988--1996, 2014.

\bibitem{deepid3}
Y.~Sun, D.~Liang, X.~Wang, and X.~Tang.
\newblock Deepid3: Face recognition with very deep neural networks.
\newblock {\em arXiv preprint arXiv:1502.00873}, 2015.

\bibitem{sun13iccv1}
Y.~Sun, X.~Wang, and X.~Tang.
\newblock Hybrid deep learning for face verification.
\newblock In {\em ICCV}, 2013.

\bibitem{DEEPID}
Y.~Sun, X.~Wang, and X.~Tang.
\newblock Deep learning face representation from predicting 10,000 classes.
\newblock In {\em Computer Vision and Pattern Recognition (CVPR), 2014 IEEE
  Conference on}, pages 1891--1898. IEEE, 2014.

\bibitem{DEEPID2P}
Y.~Sun, X.~Wang, and X.~Tang.
\newblock Deeply learned face representations are sparse, selective, and
  robust.
\newblock {\em arXiv preprint arXiv:1412.1265}, 2014.

\bibitem{szegedy15cvpr}
C.~Szegedy, W.~Liu, Y.~Jia, P.~Sermanet, S.~Reed, D.~Anguelov, D.~Erhan,
  V.~Vanhoucke, and A.~Rabinovich.
\newblock Going deeper with convolutions.
\newblock In {\em CVPR}, 2015.

\bibitem{taigman14cvpr}
Y.~Taigman, M.~Yang, M.~Ranzato, and L.~Wolf.
\newblock {DeepFace}: Closing the gap to human-level performance in face
  verification.
\newblock In {\em CVPR}, 2014.

\bibitem{deepface}
Y.~Taigman, M.~Yang, M.~Ranzato, and L.~Wolf.
\newblock Deepface: Closing the gap to human-level performance in face
  verification.
\newblock In {\em Computer Vision and Pattern Recognition (CVPR), 2014 IEEE
  Conference on}, pages 1701--1708. IEEE, 2014.

\bibitem{eigenface}
M.~A. Turk and A.~P. Pentland.
\newblock Face recognition using eigenfaces.
\newblock In {\em Computer Vision and Pattern Recognition}, pages 586--591,
  1991.

\bibitem{vincent2010stacked}
P.~Vincent, H.~Larochelle, I.~Lajoie, Y.~Bengio, and P.-A. Manzagol.
\newblock Stacked denoising autoencoders: Learning useful representations in a
  deep network with a local denoising criterion.
\newblock {\em Journal of Machine Learning Research}, 11(Dec):3371--3408, 2010.

\bibitem{weinberger09jmlr}
K.~Weinberger and L.~Saul.
\newblock Distance metric learning for large margin nearest neighbor
  classification.
\newblock {\em JMLR}, 10:207--244, 2009.

\bibitem{weinmann14eccv}
M.~Weinmann, J.~Gall, and R.~Klein.
\newblock Material classification based on training data synthesized using a
  {BTF} database.
\newblock In {\em ECCV}, 2014.

\bibitem{LSSF}
X.~Xie, W.-S. Zheng, J.~Lai, P.~C. Yuen, and C.~Y. Suen.
\newblock Normalization of face illumination based on large-and small-scale
  features.
\newblock {\em IEEE Trans. Image Processing}, 20(7):1807--1821, 2011.

\bibitem{yi15shared}
D.~Yi, Z.~Lei, and S.~Z. Li.
\newblock Shared representation learning for heterogeneous face recognition.
\newblock In {\em International Conference on Automatic Face and Gesture
  Recognition}, 2015.

\bibitem{WEBFACE}
D.~Yi, Z.~Lei, S.~Liao, and S.~Z. Li.
\newblock Learning face representation from scratch.
\newblock {\em arXiv preprint arXiv:1411.7923}, 2014.

\bibitem{yu2016sketchAnet}
Q.~Yu, Y.~Yang, F.~Liu, Y.-Z. Song, T.~Xiang, and T.~M. Hospedales.
\newblock Sketch-a-net: A deep neural network that beats humans.
\newblock {\em International Journal of Computer Vision}, 2016.

\bibitem{zeiler14eccv}
M.~Zeiler and R.~Fergus.
\newblock Visualizing and understanding convolutional networks.
\newblock In {\em ECCV}, 2014.

\bibitem{zhang14eccv}
Z.~Zhang, P.~Luo, C.~Loy, and X.~Tang.
\newblock Facial landmark detection by deep multi-task learning.
\newblock In {\em ECCV}, 2014.

\bibitem{zhu2015face}
X.~Zhu, Z.~Lei, X.~Liu, H.~Shi, and S.~Z. Li.
\newblock Face alignment across large poses: A 3d solution.
\newblock {\em arXiv preprint arXiv:1511.07212}, 2015.

\bibitem{zhu2015high}
X.~Zhu, Z.~Lei, J.~Yan, D.~Yi, and S.~Z. Li.
\newblock High-fidelity pose and expression normalization for face recognition
  in the wild.
\newblock In {\em Proceedings of the IEEE Conference on Computer Vision and
  Pattern Recognition}, pages 787--796, 2015.

\end{thebibliography}
\end{document}